\PassOptionsToPackage{capitalise}{cleveref}
\documentclass[]{fairmeta}

\usepackage{hyperref}
\usepackage{url}
\usepackage[utf8]{inputenc} %
\usepackage[T1]{fontenc}    %
\usepackage{hyperref}       %
\usepackage{url}            %
\usepackage{booktabs}       %
\usepackage{amsfonts}       %
\usepackage{nicefrac}       %
\usepackage{microtype}      %
\usepackage{xcolor}         %

\usepackage{orcidlink}
\usepackage{tikz}
\usepackage{adjustbox}
\usepackage{wrapfig}
\usepackage{multirow}
\usepackage{algorithm}
\usepackage{algorithmic}
\usepackage{wrapfig}
\usepackage{bbm}
\usepackage{tikz}
\usepackage{epigraph} 
\usepackage{threeparttable}
\usepackage{subcaption}
\usepackage{soul}
\usepackage{wrapfig}
\usepackage{amssymb}
\usepackage{amsmath}
\usepackage{xspace}

\usepackage{arydshln}
\usetikzlibrary{arrows.meta}
\usepackage{tcolorbox}
\usepackage{listings}
\usepackage{float}
\usepackage[dvipsnames]{xcolor}
\usepackage{makecell}
\usepackage{pifont}
\usepackage[table]{xcolor}
\newcommand{\cmark}{\textcolor{green!55!black}{\ding{51}}}
\newcommand{\xmark}{\textcolor{red!75!black}{\ding{55}}}
\definecolor{teasergreen}{RGB}{0,170,70}
\usepackage{capt-of}
\definecolor{stageA}{RGB}{50,130,205}   %
\definecolor{stageB}{RGB}{50,160,75}    %
\definecolor{stageC}{RGB}{225,160,0}    %

\RequirePackage{xspace}
\makeatletter
\DeclareRobustCommand\onedot{\futurelet\@let@token\@onedot}
\def\@onedot{\ifx\@let@token.\else.\null\fi\xspace}

\makeatother

\crefname{figure}{Fig.}{Fig.}

\crefname{table}{Tab.}{Tab.}

\crefname{section}{Sec.}{Sec.}

\crefname{appendix}{Appendix}{Appendices}

\title{OuroWorld: Bringing Any 3D World Alive as Diverse, Endlessly Looping 3D Cinemagraphs}

\author[1,2]{You-Zhe Xie}
\author[1]{Ting-Wei Chou}
\author[1]{Yu-Hsuan Li}
\author[2]{Kaipeng Zhang}
\author[2,\dagger]{Zhixiang Wang}
\author[1,\dagger]{Yu-Lun Liu}

\affiliation[1]{National Yang Ming Chiao Tung University}
\affiliation[2]{Alaya Lab}

\contribution[\dagger]{Corresponding authors}

\abstract{
Recent 3D world models generate photorealistic, explorable scenes that remain frozen in time. OuroWorld is a mask-free framework that turns any static 3D Gaussian Splatting scene into a 3D cinemagraph: a dynamic scene with vivid, diverse motion looping seamlessly from any viewpoint. A vision-language model infers plausible dynamics and guides a video model to synthesize a reference video, which we lift and complete into multi-view videos. To learn from this imperfect supervision, we propose Inconsistency-Robust Periodic 4DGS: a Fourier-series deformation field guarantees looping by construction, while a Grounded Drift Field anchored at the reference view absorbs cross-view inconsistency. Unlike prior Eulerian methods limited to fluid-like motion, we capture general deformation, object motion, and illumination change. We introduce a ground-truth-free evaluation covering vividness, naturalness, loop seam coherence, and scene quality. On 39 reconstructed and generated scenes, OuroWorld outperforms all baselines and wins 70.8\%--99.0\% of user-study comparisons.

\github{\url{https://ouroworld.userwei.com}}
\correspondence{\email{yulunliu@cs.nycu.edu.tw}, \email{wangzx1994@gmail.com}}
}
\date{\today}

\begin{document}
\maketitle

\begin{figure*}[h]
    \centering
    \includegraphics[width=0.85\linewidth]{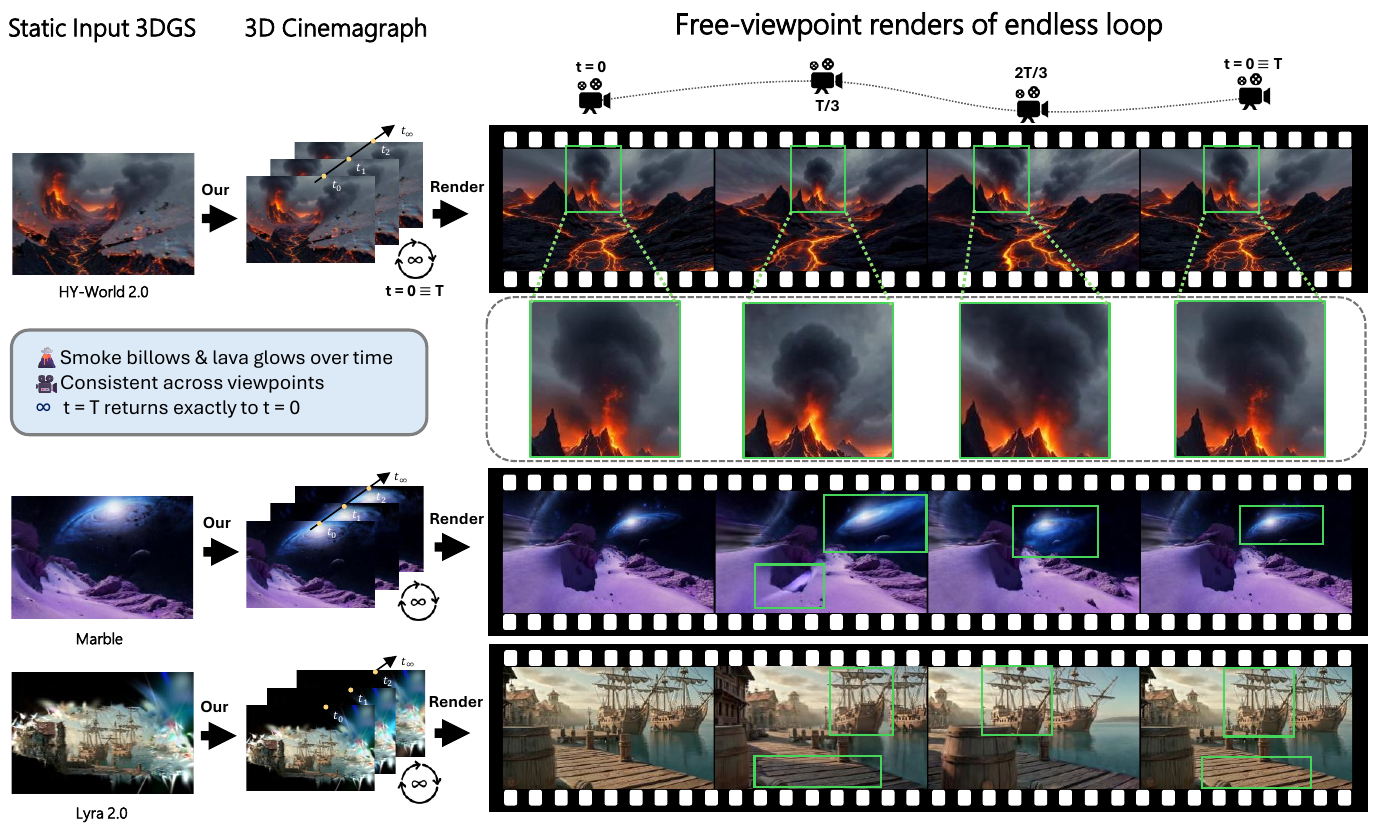}
    \caption{\textbf{OuroWorld brings static 3D worlds alive.}
    \emph{Left}: Given a static 3DGS scene from any source, e.g., the 3D world models HY-World 2.0, Marble, and Lyra 2.0, OuroWorld converts it into a 3D cinemagraph: a dynamic scene with vivid, diverse motion that loops endlessly.
   \emph{Right}: free-viewpoint renders along a moving camera, with \textcolor{teasergreen}{\textbf{green boxes}} highlighting regions with noticeable dynamics.
    Smoke billows in the volcano (top), the galaxy scatters stardust (middle), and the ships bob up and down on the water (bottom).
    Motion stays consistent across viewpoints, and the scene at $t{=}T$ returns to $t{=}0$, so the loop is seamless.
    See the supplementary videos for full results.}
    \label{figs:teaser}
\end{figure*}

\section{Introduction}
\label{sec:introduction}

\begin{figure}[t]
  \centering
  \begin{minipage}[c]{0.49\linewidth}
    \centering
    \input{figs_tex/motivation}
  \end{minipage}\hfill
  \begin{minipage}[c]{0.49\linewidth}
    \centering
    \centering
\label{tab:method_comparison}
\footnotesize
\setlength{\tabcolsep}{2.5pt}                                  %
\renewcommand{\cellset}{\renewcommand{\arraystretch}{0.9}}    %
\setlength{\aboverulesep}{3pt}                                 %
\setlength{\belowrulesep}{3pt}                                 %
\resizebox{\linewidth}{!}{%
\begin{tabular}{lccccc}
\toprule
Method
& Mask-free
& \makecell{True 3D}
& \makecell{Mathematical\\Looping}
& \makecell{Diverse\\Motion}
& Illumination \\
\midrule
Gaussians-to-Life
& \xmark & \cmark & \xmark & \cmark & \xmark \\
3D Cinemagraphy
& \xmark & \makecell{\xmark\\{\scriptsize (LDIs)}} & \cmark & \makecell{\xmark\\{\scriptsize (fluid only)}} & \xmark \\
LoopGaussian
& \xmark & \cmark & \cmark & \makecell{\xmark\\{\scriptsize (soft material only)}} & \xmark \\
3D-MOM
& \xmark & \cmark & \xmark & \makecell{\xmark\\{\scriptsize (fluid only)}} & \xmark \\
\midrule
\rowcolor{gray!12}
\textbf{Ours}
& \cmark & \cmark & \cmark & \cmark & \cmark \\
\bottomrule
\end{tabular}}

  \end{minipage}

  \begin{minipage}[t]{0.49\linewidth}
    \captionof{figure}{\textbf{Motivation.}
    \textcolor{stageA}{\emph{Left:}} 3D world models generate frozen worlds; 3D cinemagraphs bring them alive.
    \textcolor{stageC}{\emph{Middle:}} Existing 3D cinemagraph methods rely on Eulerian flow and are limited to fluid-like motion.
    \textcolor{stageB}{\emph{Right:}} We turn any static 3D scene into a 3D cinemagraph, which requires
    (1) seamless looping and (2) vivid, diverse dynamics.}
    \label{fig:motivation}
  \end{minipage}\hfill
  \begin{minipage}[t]{0.49\linewidth}
    \captionof{table}{\textbf{Capability comparison.}
    \emph{Mask-free}: no user-provided 2D/3D motion mask.
    \emph{True 3D}: free-viewpoint rendering, not pseudo-3D.
    \emph{Mathematical looping}: the 4D output is guaranteed to loop.
    \emph{Diverse motion}: not limited to one motion type.
    \emph{Illumination}: time-varying lighting.
    Ours is the only method that satisfies all five properties.}
    \label{tab:method_comparison}
  \end{minipage}
\end{figure}

Recent 3D world models, such as HY-World~2.0~\cite{hy2026hy}, Marble~\cite{worldlabs2025marble}, and Lyra~2.0~\cite{shen2026lyra}, generate photorealistic, freely explorable 3D scenes, yet these worlds are frozen in time: leaves do not sway, mist does not drift, and lights do not flicker. Our goal is to bring them alive, endlessly (\cref{figs:teaser}).

In 2D, cinemagraphs achieve this by adding seamlessly looping motion to a photograph~\cite{liao2013automated,holynski2021animating,mahapatra2026dreamloop}, but their 3D extensions remain limited. Multiplane or layered-depth methods~\cite{ma20233d,li20233d,shen2023make} are pseudo-3D with small viewpoint ranges, while LoopGaussian~\cite{li2024loopgaussian} and 3D-MOM~\cite{jin2025optimizing} use 3DGS~\cite{kerbl20233d} but rely on Eulerian flow, which captures only fluid-like motion and cannot express general deformation, object motion, or illumination change. All of them also require a motion mask and cannot decide where and how the scene should move (\cref{fig:motivation,tab:method_comparison}).

Video generation models~\cite{kong2024hunyuanvideo,wan2025wan} synthesize diverse, realistic dynamics that fill this gap, but turning them into a 3D cinemagraph poses two challenges: (1) generated videos are only approximately periodic, so fitting them leaves a visible seam; and (2) a single view cannot constrain a 4D scene, yet generated multi-view videos are mutually inconsistent, so fitting them blurs the scene and produces erratic motion.

We present OuroWorld, a mask-free framework that converts any static 3DGS scene into a 3D cinemagraph (\cref{figs:method}). A vision-language model~\cite{singh2025openai} infers plausible dynamics and guides a video model to generate a reference video; we extend this to multi-view videos and fit them with an Inconsistency-Robust Periodic 4DGS. It resolves (1) with a Fourier-series deformation that loops by construction, and (2) with a Grounded Drift Field that absorbs cross-view inconsistency while anchored at the reference view.

Since no ground truth exists for this task, we further propose a ground-truth-free evaluation of vividness, naturalness, loop seam coherence, and scene quality. On 30 world-model-generated and 9 reconstructed Mip-NeRF~360~\cite{barron2022mip} scenes, our method outperforms all baselines and is preferred in 70.8\%--99.0\% of user-study comparisons.

Our contributions are as follows:
\begin{itemize}
    \item The first mask-free 3D cinemagraph framework, turning any static 3DGS scene into a vivid, seamlessly looping dynamic scene with general motion, and illumination change beyond Eulerian methods.
    \item Inconsistency-Robust Periodic 4DGS guaranteeing looping via a Fourier-series deformation and learns robustly from inconsistent generated videos via a novel Grounded Drift Field.
    \item A perceptually-aligned ground-truth-free evaluation framework for arbitrary scenes, covering scene quality, loop seam coherence, vividness, and naturalness, under which our method outperforms all baselines on both this framework and the user study.
\end{itemize}

\section{Related Work}
\label{sec:related}

\textbf{Cinemagraphs and looping videos.}
A cinemagraph is a photograph in which selected regions move in a seamless loop.
Early systems extract such loops from an input video by finding seamless frame or per-pixel periods~\cite{schodl2023video,tompkin2011towards,yeh2012approach,liao2013automated,liao2015fast,oh2017personalized}, optionally de-animating static regions~\cite{bai2012selectively}.
Animating a single photograph instead requires a motion prior: stochastic motion textures~\cite{chuang2005animating}, class-specific physical models~\cite{jhou2015animating}, detected periodic patterns~\cite{halperin2021endless}, learned time-lapse~\cite{endo2019animating} or Eulerian-flow predictors~\cite{holynski2021animating,mahapatra2022controllable,fan2023simulating,mahapatra2023text,choi2024stylecinegan}, and frequency-domain oscillations~\cite{li2024generative}.
Image-to-video diffusion models~\cite{blattmann2023stable,xing2024dynamicrafter} now provide far richer priors, and loop-specific variants cyclically shift latents or positional embeddings~\cite{bi2025mobius,dong2026loopy}, animate salient objects~\cite{wang2024loopanimate}, or condition the first and last frames on the same photograph~\cite{mahapatra2026dreamloop}, which we adopt to obtain a reference loop.
All of these, however, animate a \emph{2D} image plane, and the Eulerian ones are confined to fluid-like motion.
We instead produce a \emph{3D} scene renderable from arbitrary viewpoints, with loop closure enforced by the representation itself rather than by the 2D generator.

\textbf{Static 3D scene representations and generated worlds.}
Single-image view synthesis lifts a photo into a shallow proxy---a point cloud~\cite{niklaus20193d,wiles2020synsin}, multiplane image~\cite{zhou2018stereo,tucker2020single}, or layered depth image~\cite{shade1998layered,tulsiani2018layer,shih20203d}---with limited parallax, whereas neural radiance fields~\cite{mildenhall2021nerf,barron2022mip} and 3D Gaussian Splatting (3DGS)~\cite{kerbl20233d,yu2024mip,huang20242d} reconstruct complete scenes, the latter in real time. They assume \emph{consistent} captured observations, and some work handles inconsistency with per-image latent codes~\cite{martin2021nerf,park2021nerfies,kulhanek2024wildgaussians}, uncertainty-based transient removal~\cite{hou20253d}, view-dependent illumination compensation~\cite{yang2026ic}, or ICP-estimated per-frame drift~\cite{hollein2026world}.
3DGS is also the output of scene generators that build explorable worlds by progressive inpainting and lifting~\cite{hollein2023text2room,li2024genrc,chung2023luciddreamer,yu2024wonderjourney,yu2025wonderworld,zhou2024dreamscene360,huang2026vlm,liang2025wonderland}, by iteratively refining renders with diffusion~\cite{lee2026skyfall}, or with large-scale world models~\cite{team2025hunyuanworld,hy2026hy,huang2025voyager,chen2026pantheon360,worldlabs2025marble,bahmani2026lyra,shen2026lyra}.

\textbf{Dynamic 3D representations and 4D generation.}
Dynamic radiance fields and 3DGS add time via canonical-space deformation~\cite{pumarola2021d,park2021nerfies,park2021hypernerf,yang2024deformable,wu20244d,fan2025spectromotion}, scene flow~\cite{li2021neural}, factorized space-time grids~\cite{fridovich2023k,cao2023hexplane,wang2022fourier}, per-frame optimization~\cite{luiten2024dynamic}, sparse control points~\cite{huang2024sc}, native 4D primitives~\cite{yang2024real,duan20244d}, or per-Gaussian temporal bases~\cite{li2024spacetime,lin2024gaussian,kratimenos2024dynmf,chen2026periodic,shih2024modeling,zhang2026fast,chan2026adagar}, and now handle casual monocular videos~\cite{liu2023robust,wang2025shape,lei2025mosca,shih2025prior}, even freezing them into a static scene~\cite{chien2026splannequin}.
4D generation synthesizes objects via score distillation~\cite{singer2023text,bahmani20244d,ling2024align,ren2023dreamgaussian4d,bahmani2024tc4d} or multi-view video diffusion~\cite{xie2025sv4d,ren2024l4gm,jiang2024animate3d}, and recently whole scenes~\cite{yu20244real,wang20254real,sun2024dimensionx,zhao2025genxd,wu2025cat4d,liu2025free4d,pan2025diff4splat}, with Free4D~\cite{liu2025free4d} supervising generated views with diffusion-refined renders as we do.
We restrict deformation to a Fourier series, making motion \emph{periodic} by construction.
Whereas per-image appearance embeddings and illumination compensation~\cite{martin2021nerf,kulhanek2024wildgaussians,yang2026ic} explain photometric variation across captured images, our Grounded Drift Field absorbs \emph{generative} inconsistency as a view-conditioned drift, anchored at the reference view and discarded at inference.

\textbf{3D cinemagraphs and scene animation.}
3D cinemagraphs splat single-image 2D Eulerian flow on layered depth~\cite{li20233d,shen2023make}, assemble loops from asynchronous multi-view captures~\cite{ma20233d}, or move 3D Gaussians along Eulerian fields~\cite{li2024loopgaussian,jin2025optimizing}; all rely on a user- or network-provided \emph{mask} and stationary flow.
Scene animation instead borrows motion from video diffusion~\cite{lee2024vividdream,wimmer2025gaussians}, physics simulation~\cite{xie2024physgaussian,zhang2024physdreamer,huang2025dreamphysics,liu2024physics3d}, or language-model-written motion fields~\cite{kiray2026promptvfx}, mostly within user-specified regions or, as in the concurrent AniGS~\cite{cheng2026anigs}, segmentation masks; none targets a seamless loop.
Our method is \emph{mask-free}: a vision-language model describes the motion, a video diffusion model realizes it, and a periodic 4D representation guarantees an endless loop from any viewpoint (Table~\ref{tab:method_comparison}).

\section{Method}
\label{sec:method}

\begin{figure*}[t]
    \centering
    \includegraphics[width=0.8\linewidth]{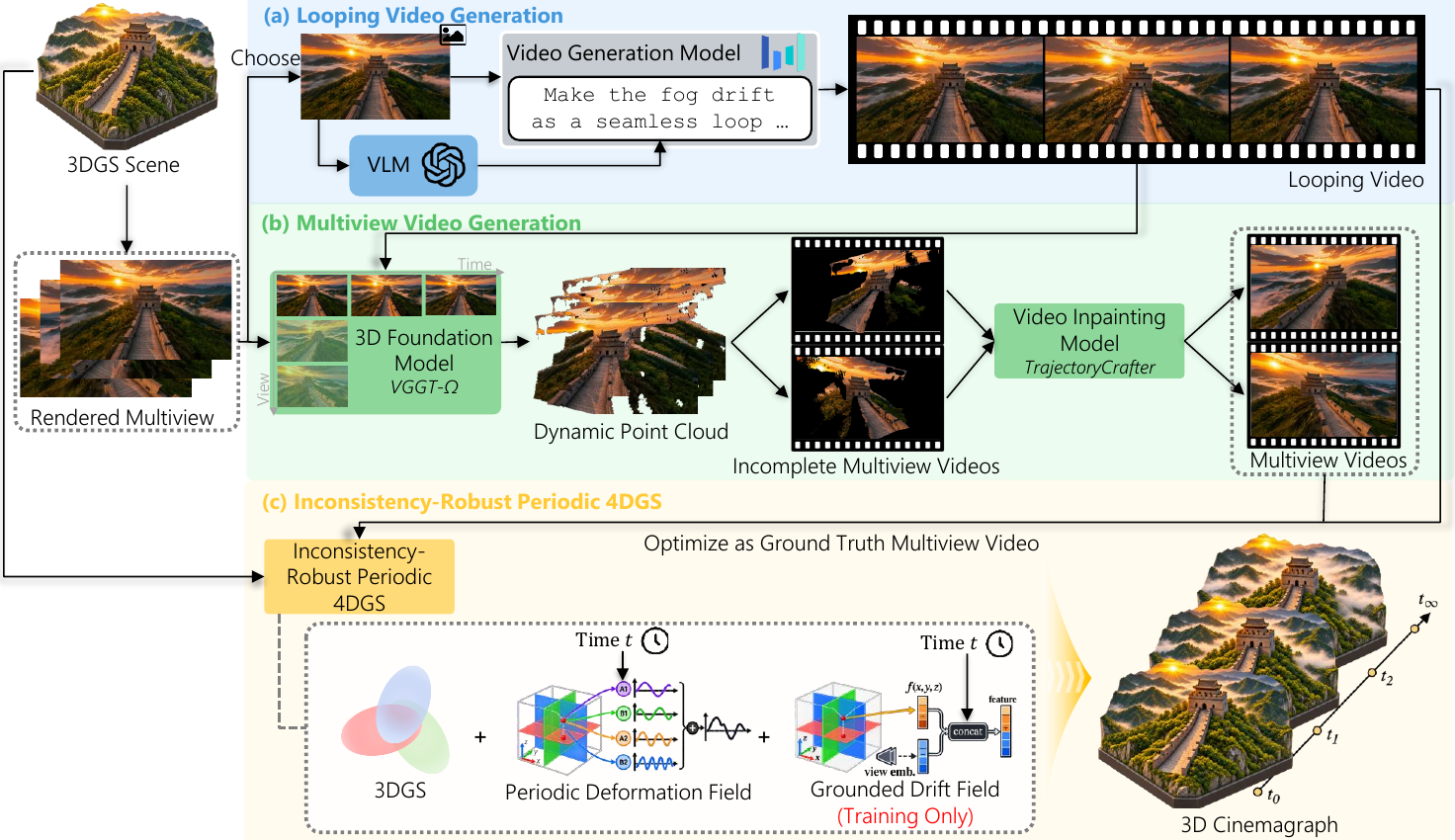}

    \caption{\textbf{Overview of OuroWorld.}
    \textcolor{stageA}{\textbf{(a) Looping video generation.}} A VLM selects a reference view rendered from the input 3DGS, infers plausible scene dynamics, and prompts a video generation model to synthesize an approximately looping reference video.
    \textcolor{stageB}{\textbf{(b) Multi-view video generation.}} A 3D foundation model (VGGT-$\Omega$) lifts the reference video into a dynamic point cloud, using static multi-view renders to improve depth. The point cloud is rendered from $N$ surrounding viewpoints into incomplete videos with disoccluded holes, which a video inpainting model (TrajectoryCrafter) completes into multi-view videos.
    \textcolor{stageC}{\textbf{(c) Inconsistency-Robust Periodic 4DGS.}} The 3D cinemagraph is represented as a canonical 3DGS deformed by a Fourier-parameterized Periodic Deformation Field, which loops by construction. A Grounded Drift Field absorbs cross-view inconsistency of the generated videos during training and is discarded at inference.}

    \label{figs:method}
\end{figure*}

Given a static 3D Gaussian Splatting (3DGS)~\cite{kerbl20233d} scene from any source, we automatically convert it into a 3D cinemagraph.
There are three challenges: (1) obtaining diverse dynamics, (2) guaranteeing looping motion, and (3) learning 4D from inconsistent generated multi-view supervision.
We propose \textbf{OuroWorld}, a 3-stage framework (\cref{figs:method}) including looping video generation (\cref{sec:loop_video}), multi-view video generation (\cref{sec:mv_video}), and Inconsistency-Robust Periodic 4DGS optimization (\crefrange{sec:irp4dgs}{sec:optim}). See~\cref{suppl_sec:method_detail} for details.

\vspace{-1em}
\subsection{Looping Video Generation}
\label{sec:loop_video}

Unlike Eulerian flow methods~\cite{holynski2021animating,li20233d,jin2025optimizing,li2024loopgaussian}, which are limited to fluid-like motion and require manual masks, modern video diffusion models~\cite{kong2024hunyuanvideo,wan2025wan} synthesize diverse, realistic dynamics, including approximately looping videos.
As shown in \cref{figs:method}a, given a reference image $I_{\text{ref}}$ rendered from the input 3DGS, a vision-language model (VLM; GPT-5.5~\cite{singh2025openai}) infers what should move and how, and writes a prompt describing natural cyclic motion (details in~\cref{suppl_sec:method_detail}).
Conditioned on this, a video generation model (Seedance~2.0~\cite{seedance2026seedance}) produces a 10-second reference video $V_{\text{ref}}$ with $I_{\text{ref}}$ as both first and last frame, and the VLM filters out results with unnatural motion.
This module accepts any video generator capable of producing approximately looping videos, including prompt-controllable ones~\cite{bi2025mobius,mahapatra2026dreamloop}.

\vspace{-1em}
\subsection{Multi-view Video Generation}
\label{sec:mv_video}

A single view is insufficient to constrain a full 4D scene. We therefore extend the reference video into multi-view videos. 
As shown in \cref{figs:method}b, we feed the reference video together with multi-view renders from input 3DGS into a 3D foundation model (VGGT-$\Omega$~\cite{wang2026vggt,wang2025vggt}), to estimate a per-frame dynamic point cloud of the reference video~\cite{wang2024dust3r,zhang2025monst3r}; the static renders serve only to improve depth estimation. We then sample $N{=}20$ viewpoints uniformly within $\pm 20^\circ$ of horizontal rotation around a pivot and render the dynamic point cloud from these viewpoints into incomplete videos containing disoccluded holes. Finally, the video inpainting model (TrajectoryCrafter~\cite{yu2025trajectorycrafter}), conditioned on the reference video, inpaints each view, producing multi-view videos $\{V_v\}_{v=1}^{N}$.

\vspace{-1em}
\subsection{Inconsistency-Robust Periodic 4DGS}
\label{sec:irp4dgs}

As shown in \cref{figs:method}c, we represent the 3D cinemagraph with Inconsistency-Robust Periodic 4DGS, comprising a canonical 3DGS $\mathcal{G}$, a Periodic Deformation Field $\mathcal{P}$, and a Grounded Drift Field $\Delta$. Each Gaussian in $\mathcal{G}$ is parameterized by position $\mathbf{x}\in\mathbb{R}^3$, rotation $\mathbf{q}$, scale $\mathbf{s}$, opacity $\sigma$, and color $\mathbf{c}$.

\vspace{-1em}
\subsubsection{Periodic Deformation Field}
\label{sec:periodic}

We parameterize the temporal variation of each Gaussian with a Fourier series (\cref{figs:method}c), which is periodic by construction and can approximate arbitrary smooth periodic motion, enabling diverse dynamics beyond the fluid-like motion of Eulerian flow.

Following factorized spatial encodings~\cite{chan2022efficient,fridovich2023k,cao2023hexplane,wu20244d}, we query a triplane at the canonical center $\mathbf{x}$ and obtain a feature $\mathbf{F}(\mathbf{x})\in\mathbb{R}^{(2K+1)C}$, which we split into per-frequency coefficients $\mathbf{a}_0,\{\mathbf{a}_k,\mathbf{b}_k\}_{k=1}^{K}\in\mathbb{R}^{C}$. The time-dependent feature is
{\small
\begin{equation}
  \mathbf{f}(\mathbf{x},t) = \mathbf{a}_0 + \sum_{k=1}^{K}\Big[\mathbf{a}_k\cos\!\Big(\frac{2\pi k t}{T}\Big) + \mathbf{b}_k\sin\!\Big(\frac{2\pi k t}{T}\Big)\Big],
  \label{eq:fourier}
\end{equation}
}
where $T$ is the loop period and $K$ the temporal bandwidth.
An MLP decodes it into attribute offsets $\mathcal{P}(\mathbf{x},t)=(\delta\mathbf{x},\delta\mathbf{q},\delta\mathbf{s},\delta\mathbf{c})=\mathrm{MLP}_{\mathcal{P}}(\mathbf{f}(\mathbf{x},t))$.
Since $\mathbf{f}(\mathbf{x},t{+}T)=\mathbf{f}(\mathbf{x},t)$, $\mathcal{P}$ is $T$-periodic and $\mathcal{G}+\mathcal{P}(\mathbf{x},t)$ loops seamlessly regardless of the optimization outcome.
Unlike prior Fourier-domain dynamic representations~\cite{lin2024gaussian,li2024spacetime}, which use such bases for fitting capacity, we exploit periodicity to guarantee looping.

\vspace{-1em}
\subsubsection{Grounded Drift Field}
\label{sec:drift}

\begin{wrapfigure}{r}{0.5\textwidth}
    \centering
    \includegraphics[width=\linewidth]{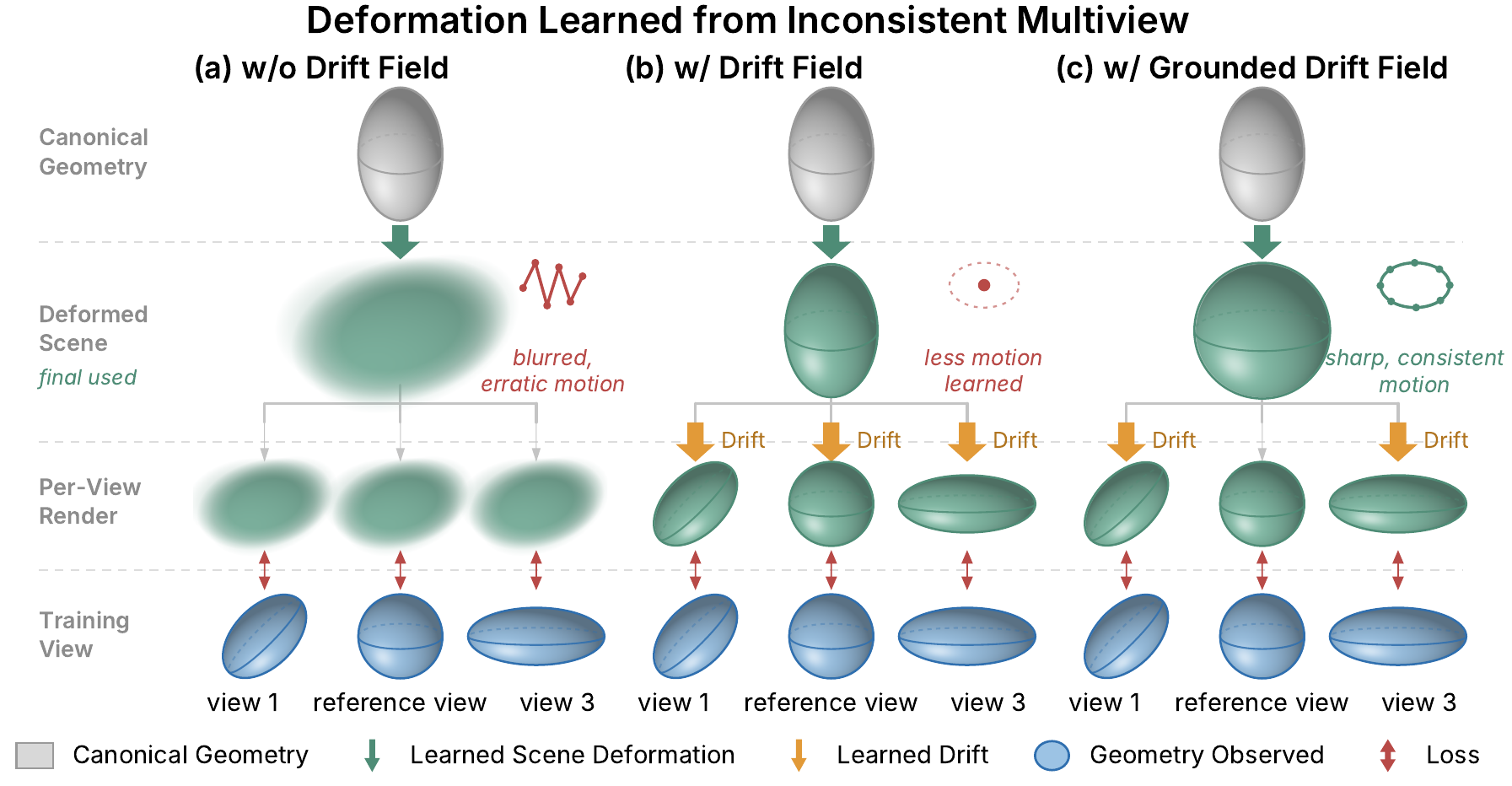}
    \caption{\textbf{Learning deformation from inconsistent multi-view videos.}
    Rows: canonical geometry, deformed scene at inference, per-view training renders, and generated training views.
    \textbf{(a) w/o Drift Field:} conflicting views are averaged, causing blur and erratic motion.
    \textbf{(b) w/ Drift Field:} drift on every view absorbs too much of the dynamics and is discarded at inference, reducing motion.
    \textbf{(c) w/ Grounded Drift Field (ours):} with no drift at the reference view, the deformation field must reproduce the reference dynamics, yielding a sharp scene with consistent motion.}
    \label{figs:drift_flow_demo}
\end{wrapfigure}

\textbf{Drift.} The generated multi-view videos are inconsistent across views. Fitting them directly forces the deformation field to average conflicting observations, blurring the scene and producing erratic motion (Fig.~\ref{figs:drift_flow_demo}a). Video-to-World~\cite{hollein2026world} explains such inconsistency as per-view \emph{generative drift}, estimates it with ICP~\cite{besl1992method}, and adds the drift before computing the loss, so the drift is not baked into the scene. However, ICP only aligns geometry, missing drift in attributes such as \emph{scale} and \emph{color}, and assumes a static scene, failing on dynamic content.

We instead \emph{learn} drift jointly with the 4D scene. A drift field $\Delta$, sharing the triplane-plus-MLP architecture of $\mathcal{P}$, predicts view-specific, time-varying offsets to position, rotation, \emph{scale} and  \emph{color},
$\Delta_v(\mathbf{x},t) = \mathrm{MLP}_{\Delta}\big(\mathbf{F}_\Delta(\mathbf{x}), \gamma(t), \mathbf{e}_v\big)$,
from a drift spatial feature $\mathbf{F}_\Delta(\mathbf{x})$, a time embedding $\gamma(t)$, and a per-view embedding $\mathbf{e}_v$. Adding $\Delta_v$ when rendering view $v$ lets the loss attribute view-specific inconsistency to $\Delta_v$ rather than the shared scene. At inference, we discard $\Delta$ and render $\mathcal{G}+\mathcal{P}(\mathbf{x},t)$, which retains dynamics that are largely free of cross-view inconsistency.

\textbf{Grounding.}
Learning drift in dynamic scenes introduces an ambiguity: since both $\mathcal{P}$ and $\Delta$ depend on time, drift applied to all views can more easily explain most dynamics, leaving the final scene less dynamic (Fig.~\ref{figs:drift_flow_demo}b). We resolve this by grounding the optimization in the reference view, whose dynamics come directly from the video generation model and are the most reliable:
{\small
\begin{equation}
  \mathcal{G}_{t,v} =
  \begin{cases}
    \mathcal{G} + \mathcal{P}(\mathbf{x},t), & v = \text{ref},\\[2pt]
    \mathcal{G} + \mathcal{P}(\mathbf{x},t) + \Delta_v(\mathbf{x},t), & \text{otherwise}.
  \end{cases}
  \label{eq:grounded}
\end{equation}
}
$\mathcal{P}$ must thus reproduce the reference dynamics by itself, while $\Delta_v$ only absorbs each view's residual deviation from the reference (Fig.~\ref{figs:drift_flow_demo}c).

\vspace{-1em}
\subsection{Scene-View Consistent Optimization}
\label{sec:optim}

While the Grounded Drift Field handles inconsistencies among generated views, the generated content may also be inconsistent with the input 3DGS. Following Free4D~\cite{liu2025free4d}, we supervise highly consistent images, namely the reference video $V_{\text{ref}}$ and the $t{=}0$ multi-view renders of the input 3DGS, with an $\ell_1$ loss $\mathcal{L}_1$. For the remaining generated frames, we perturb the rendering with noise and denoise it with an image diffusion model (Stable Diffusion v1.5~\cite{rombach2022high}) guided by the generated frame, yielding refined images. We then supervise the rendering with an LPIPS loss~\cite{zhang2018unreasonable} $\mathcal{L}_{\text{p}}$ against the refined image. The total objective is $\mathcal{L} = \lambda_1\mathcal{L}_1 + \lambda_{\text{p}}\mathcal{L}_{\text{p}}$; details are provided in the~\cref{suppl_sec:optim_detail}.

\section{Experiments}
\label{sec:experiments}

We compare our method with three 3D cinemagraph methods~\cite{li20233d,li2024loopgaussian,jin2025optimizing} and one method that brings 3D scenes to life with video priors~\cite{wimmer2025gaussians}. The evaluation uses ground-truth-free automatic metrics we proposed together with a user study.

\vspace{-1em}
\subsection{Dataset}
\label{sec:exp_dataset}

We evaluate on 39 scenes, each consisting of a static 3DGS and a designated reference camera pose. Since reconstructed and generated 3DGS differ substantially in Gaussian density, geometric fidelity, and layout, we draw 9 scenes from Mip-NeRF~360~\cite{barron2022mip} and 10 each from HY-World~2.0~\cite{hy2026hy}, Marble~\cite{worldlabs2025marble}, and Lyra~2.0~\cite{shen2026lyra} (details in~\cref{suppl_sec:dataset_detail}). The scenes span indoor rooms, landscapes, architecture, and stylized environments, covering both the fluid and cloth-like dynamics targeted by Eulerian methods and motions beyond them, such as flickering light and twinkling scattered stardust.

\vspace{-1em}
\subsection{Baselines}
\label{sec:exp_baselines}

We compare against four methods. Gaussians-to-Life~\cite{wimmer2025gaussians} lifts video-diffusion motion~\cite{blattmann2023stable} to 3D point flow that drives a 3DGS, but does not target looping. 3D Cinemagraphy~\cite{li20233d} and 3D-MOM~\cite{jin2025optimizing} animate a single image with an Eulerian motion prior trained on fluid videos~\cite{holynski2021animating}, rendered via a layered depth image~\cite{shade1998layered} and 4D Gaussians~\cite{wu20244d}, respectively; the latter's optimized 4D output is not guaranteed to loop. LoopGaussian~\cite{li2024loopgaussian} heuristically moves only Gaussian positions along a bidirectionally blended Eulerian field. For fairness, we also apply the same GPT-5.5 filtering to all baselines. Full descriptions are in \cref{suppl_sec:baseline_detail}.

\textbf{Input alignment.} All baselines require inputs beyond a static 3DGS,
which we derive from our pipeline for a fair comparison. \emph{Input images} are rendered from the input 3DGS at the reference pose; \emph{text prompts} are the GPT-5.5~\cite{singh2025openai} dynamics prompts from \cref{sec:loop_video}; for \emph{motion masks}, a rule-based method names the objects to move and SAM~3~\cite{carion2026sam} segments them into 2D masks, which we back-project with the rendered 3DGS depth for methods requiring 3D masks. Further details are in \cref{suppl_sec:baseline_detail}.

\vspace{-1em}
\subsection{Evaluation Protocol}
\label{sec:exp_protocol}

Since all methods output scene representations, we render each under identical camera settings: a 20-second, 30~FPS video (600 frames) at $672\times384$ along two trajectories per scene. The \textbf{static camera} stays fixed at the reference pose. The \textbf{orbit camera} evaluates 3D consistency: it faces a pivot, obtained by back-projecting a selected pixel at the designated pose, at $0.95\times$ the camera-to-pivot distance, and sweeps in yaw through $0^\circ \!\rightarrow\! 20^\circ \!\rightarrow\! {-20^\circ} \!\rightarrow\! 0^\circ$ over 600 uniformly interpolated views. Both trajectories span many motion periods. We keep each method's native period and repeat it to fill 20 seconds, rather than normalizing periods, since changing the period alters motion speed and introduces artifacts unrelated to the method (\cref{suppl_sec:baseline_detail}).

\vspace{-1em}
\subsection{Evaluation Metrics}
\label{sec:exp_metrics}

Prior 3D cinemagraph works measure quality against ground truth. Their data, however, is limited to fluid datasets~\cite{holynski2021animating} or object-level synthetic scenes~\cite{li2024loopgaussian}, and does not generalize to the diverse scenes. We therefore propose a ground-truth-free evaluation framework based on properties a good 3D cinemagraph should have: \emph{vivid}, \emph{natural}, \emph{seamless loop}, \emph{high visual quality}. These metrics are perceptually aligned, as supported by the user study (\cref{sec:exp_user}).

\textbf{Vividness.} This metric measures how rich the displayed dynamics are. We temporally downsample each video to 3~FPS and define the Vividness Degree as the average of three components:
{\small
\begin{equation}
\text{Vividness} = \tfrac{1}{3}\big(\text{MV} + \text{IV} + \text{VV}\big).
\label{eq:vividness}
\end{equation}
}

Each component is computed between adjacent frames and averaged over the video, with thresholds set to the level of change perceptible to human vision:
\emph{Motion Variation} (MV) is the fraction of pixels whose optical-flow magnitude, estimated by SEA-RAFT~\cite{wang2024sea}, exceeds $0.3\%\times\min(H,W)$;
\emph{Illumination Variation} (IV) is the fraction of pixels whose grayscale luminance changes by more than 20\%;
and \emph{Visual Variation} (VV) measures global appearance change as $1-\mathrm{SSIM}$~\cite{wang2004image}, following VideoScore~\cite{he2024videoscore}.

\textbf{Naturalness.} We treat a large video generation model as an approximation of the natural video distribution and measure KVD~\cite{unterthiner2018towards} against its outputs. Using the open-source MiniMax~H3~\cite{minimax2026h3}, distinct from our Seedance~2.0 to avoid bias, we generate 20 videos per scene (780 in total), conditioned on the reference view and the GPT-5.5 dynamics prompt. For each method, we compute KVD between its renders across all scenes and both trajectories.

\textbf{Loop Seam Coherence.} We evaluate whether each cycle returns to its starting state and whether the transition across cycles is continuous, using two metrics:
\begin{itemize}
    \item \emph{Motion-Aware Loop Fidelity (MALF).} Since a static scene trivially loops, MALF rewards matching the same phase one period later relative to other phases, and is zero for a static video. With $M{=}10$ time points $\{t_i\}_{i=1}^M$ sampled in one period,
    {\small
    \begin{equation}
    \mathrm{MALF} = \frac{1}{M}\sum_{i=1}^{M}\Big[\mathrm{SSIM}\big(I_{t_i}, I_{t_i+T}\big) - \frac{1}{M-1}\sum_{j\neq i}\mathrm{SSIM}\big(I_{t_i}, I_{t_j+T}\big)\Big].
    \label{eq:MALF}
    \end{equation}
    }
    \item \emph{Seam SSIM.} The SSIM between the last frame of one cycle and the first frame of the next.
\end{itemize}
Both are computed on a single seam, making them independent of the number of cycles, and under a static camera to exclude viewpoint changes.

\textbf{Scene Quality.} We adopt VBench~\cite{huang2024vbench} \emph{Aesthetic Quality}~\cite{schuhmann2022laion} and \emph{Overall Consistency}~\cite{wang2024internvid}.

\vspace{-1em}
\subsection{Implementation Details}
\label{sec:exp_impl}

The hyperparameters of ours are listed in \cref{suppl_sec:method_detail}. For baselines, we use the official implementations (details in~\cref{suppl_sec:baseline_detail}). All experiments are run on a single NVIDIA RTX~4090 GPU.

\vspace{-1em}
\subsection{Quantitative Results}
\label{sec:exp_quant}

\begin{table}[t]
\centering
\caption{\textbf{Quantitative result.}
\emph{Vividness Degree} measures motion, illumination, and appearance change.
\emph{KVD} measures the distance to the natural video distribution.
\emph{MALF} measures whether the scene loops while it actually moves, and \emph{Seam SSIM} measures the similarity at the seam of consecutive cycles;
\emph{Aesthetic Quality} and \emph{Overall Consistency} follow VBench to assess scene quality.
\textbf{Bold} and \underline{underline} denote the best and second-best results.
Ours achieves the best Vividness, Naturalness, and Aesthetic Quality.
Near-static methods (LoopGaussian, 3D Cinemagraphy) trivially reach a high Seam SSIM but near-zero MALF, whereas ours achieves the best MALF with a high Seam SSIM.}
\label{tab:quantitative}
\setlength{\tabcolsep}{4pt}
\resizebox{0.75\textwidth}{!}{%
\begin{tabular}{clcccccc}
\toprule
& & Vividness & Naturalness
& \multicolumn{2}{c}{Loop Seam Coherence}
& \multicolumn{2}{c}{Scene Quality} \\
\cmidrule(lr){3-3} \cmidrule(lr){4-4} \cmidrule(lr){5-6} \cmidrule(lr){7-8}
& Method
& \makecell{Vividness\\Degree $\uparrow$}
& \makecell{KVD $\downarrow$}
& \makecell{Motion-Aware\\Loop Fidelity $\uparrow$}
& \makecell{Seam\\SSIM $\uparrow$}
& \makecell{Aesthetic\\Quality $\uparrow$}
& \makecell{Overall\\Consist. $\uparrow$} \\
\midrule
& Gaussians-to-Life
& \underline{0.0374} & \underline{127.47} & 0.0265 & 0.9388 & 0.6602 & 0.2109 \\
& 3D Cinemagraphy
& 0.0066 & 142.42 & 0.0134 & \underline{0.9995} & 0.6374 & \underline{0.2110} \\
& LoopGaussian
& 0.0024 & 154.65 & 0.0025 & \textbf{0.9998} & \underline{0.6718} & 0.2100 \\
& 3D-MOM
& 0.0247 & 155.23 & \underline{0.0317} & 0.9158 & 0.6245 & 0.2069 \\
\rowcolor{gray!12}
\cellcolor{white}\multirow{-5}{*}{\rotatebox{90}{Static}}
& \textbf{Ours}
& \textbf{0.0423} & \textbf{108.23} & \textbf{0.0809} & 0.9965 & \textbf{0.6748} & \textbf{0.2111} \\
\midrule
& Gaussians-to-Life
& \underline{0.5097} & \underline{51.86} & -- & -- & \underline{0.6482} & 0.2163 \\
& 3D Cinemagraphy
& 0.4945 & 102.58 & -- & -- & 0.5539 & 0.2156 \\
& LoopGaussian
& 0.5087 & 53.38 & -- & -- & 0.6470 & 0.2161 \\
& 3D-MOM
& 0.4724 & 77.83 & -- & -- & 0.5501 & \textbf{0.2191} \\
\rowcolor{gray!12}
\cellcolor{white}\multirow{-5}{*}{\rotatebox{90}{Orbit}}
& \textbf{Ours}
& \textbf{0.5117} & \textbf{50.11} & -- & -- & \textbf{0.6659} & \underline{0.2185} \\
\bottomrule
\end{tabular}%
}
\end{table}

As shown in \cref{tab:quantitative}, our method achieves the best Vividness, Naturalness, MALF, and Aesthetic Quality under both camera settings. LoopGaussian and 3D Cinemagraphy attain slightly higher Seam SSIM, but they are nearly static, as their low Vividness and MALF indicate, so their seams are trivially continuous. Our method, in contrast, combines high Seam SSIM with the highest MALF, showing that its motion is both substantial and seamlessly looping.

\vspace{-1em}
\subsection{Qualitative Results}
\label{sec:exp_qual}

\begin{figure}[t]
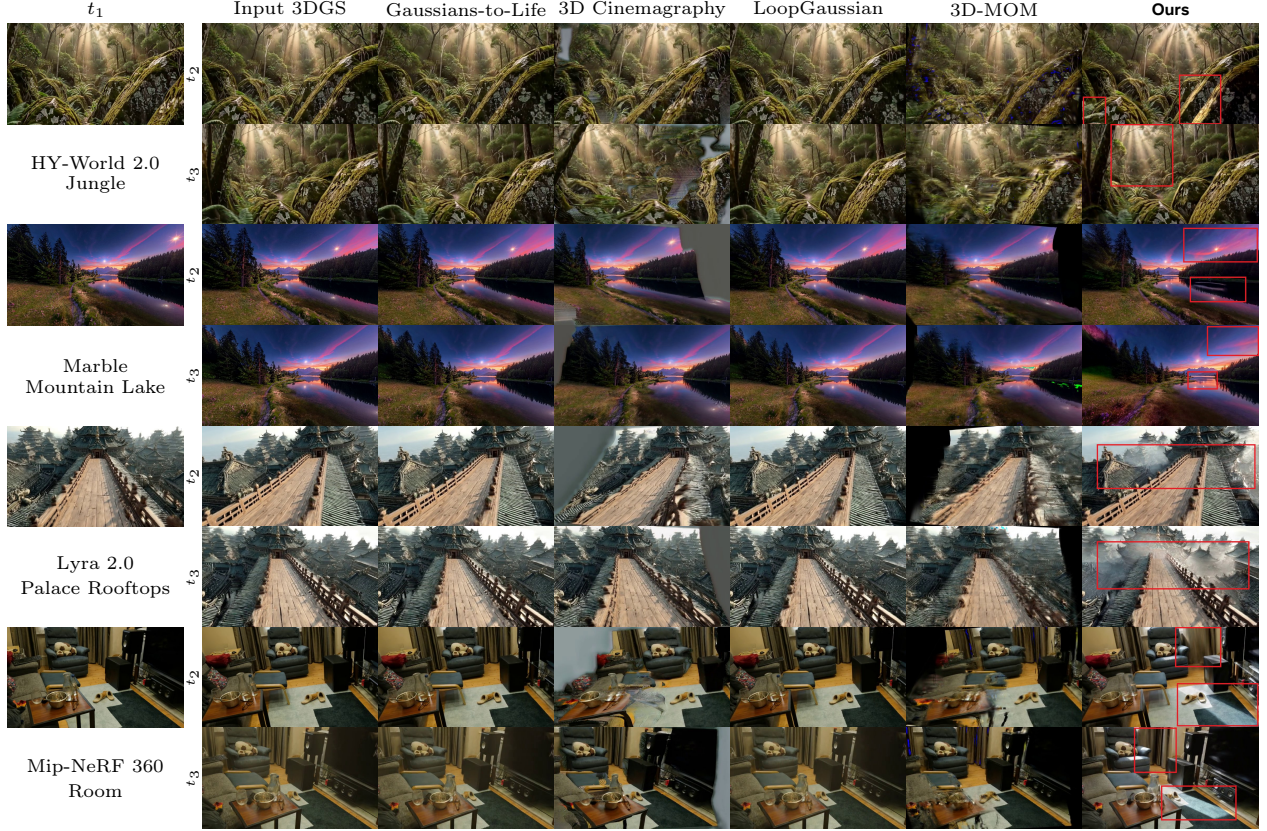

\centering
\ifdefined\qw\else\newlength{\qw}\fi
\ifdefined\qh\else\newlength{\qh}\fi
\setlength{\qw}{\dimexpr(\textwidth-0.7em-1pt)/7\relax} %
\setlength{\qh}{0.5714\qw}                             %
\newcommand{\qimg}[3]{\includegraphics[width=\qw,height=\qh]{figs/qual/#1/#2_t#3}}
\newcommand{\qfirst}[2]{\begin{tabular}{@{}c@{}}\qimg{#1}{3dgs}{1}\\
  \parbox[c][\qh][c]{\qw}{\centering\scriptsize\shortstack[c]{#2}}\end{tabular}}
\newcommand{\qstack}[2]{\begin{tabular}{@{}c@{}}\qimg{#1}{#2}{2}\\\qimg{#1}{#2}{3}\end{tabular}}
\newcommand{\qt}[1]{\rotatebox{90}{\parbox[c][0.7em][c]{\qh}{\centering\tiny $t_{#1}$}}}
\newcommand{\qtime}{\begin{tabular}{@{}c@{}}\qt{2}\\\qt{3}\end{tabular}}
\newcommand{\qhead}[1]{\parbox[c][2.4ex][c]{\qw}{\centering\scriptsize #1}}
\newcommand{\qrow}[2]{\qfirst{#1}{#2} & \qtime
  & \qstack{#1}{3dgs}
  & \qstack{#1}{g2l} & \qstack{#1}{3dc} & \qstack{#1}{loopg}
  & \qstack{#1}{3dmom} & \qstack{#1}{ours}}
\renewcommand{\arraystretch}{0}
\begin{tabular}{@{}*{8}{c@{}}}
\qhead{$t_1$} &
  & \qhead{Input 3DGS}
  & \qhead{Gaussians-to-Life} & \qhead{3D Cinemagraphy} & \qhead{LoopGaussian}
  & \qhead{3D-MOM} & \qhead{\textbf{Ours}} \\
\qrow{jungle}{HY-World 2.0 \\ Jungle} \\
\qrow{mountain_lake}{Marble \\ Mountain Lake} \\
\qrow{palace_rooftops}{Lyra 2.0 \\ Palace Rooftops} \\
\qrow{room}{Mip-NeRF 360 \\ Room} \\
\end{tabular}
\caption{\textbf{Qualitative comparison under the orbit camera}, one scene per source. The leftmost column shows the input 3DGS at $t_1$; the others show each method at later time steps $t_2$ and $t_3$.
3D Cinemagraphy and 3D-MOM produce disocclusion holes and color artifacts, while Gaussians-to-Life and LoopGaussian remain nearly static.
OuroWorld produces vivid, view-consistent dynamics such as changing illumination, rippling water, drifting fog, and curtains swaying in the wind. See the supplementary videos for full loops.}
\label{figs:qualitative}
\end{figure}

\cref{figs:qualitative} shows results under the orbit camera. 3D Cinemagraphy and 3D-MOM produce disoccluded holes at novel views, while Gaussians-to-Life and LoopGaussian remain nearly static. In contrast, our method produces vivid dynamics and stays consistent across viewpoints.

\vspace{-1em}
\subsection{User Study}
\label{sec:exp_user}

\begin{figure*}[t]
    \centering
    \includegraphics[width=1.0\linewidth]{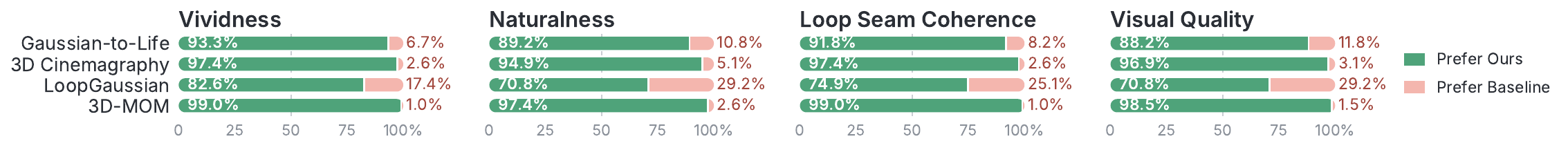}
    \caption{
    \textbf{User study.} In a two-alternative forced-choice study, participants compare OuroWorld with each baseline under four criteria.
    Bars show the percentage of votes preferring ours (left) versus the baseline (right).
    Ours is preferred in 70.8\%--99.0\% of comparisons across all baselines and criteria, with the largest margins in Vividness.
    30 participants took part; see \cref{suppl_sec:user_study_detail} for details.}
    \label{figs:user_study}
\end{figure*}

Since automatic metrics only approximate perceptual qualities, we further conduct a two-alternative forced-choice study in which 30 participants compare our result with each baseline under four criteria: Vividness, Naturalness, Loop Seam Coherence, and Visual Quality (details in \cref{suppl_sec:user_study_detail}). As shown in \cref{figs:user_study}, our method is consistently preferred across all baselines and criteria, with the largest margins in Vividness, highlighting the diverse dynamics enabled by our video generation prior. These preferences are consistent with the results in \cref{tab:quantitative} and support the validity of our designed metrics.

\vspace{-1em}
\subsection{Ablation Study}
\label{sec:exp_ablation}

\begin{table}[t]
\centering
\caption{\textbf{Ablation on consistency components.} As illustrated in \cref{figs:drift_flow_demo}, removing the \emph{drift field} blurs the scene (low VoL) and lowers both subject and background consistency; removing \emph{grounding} lets drift absorb the dynamics, reducing the scene motion (lowest Vividness); removing \emph{scene-view consistent optimization} breaks scene-view consistency and blurs the scene most severely (lowest subject consistency, VoL $-22\%$ under orbit). Our full model is best on all orbit-camera metrics.}
\label{tab:ablation_consistency}
\resizebox{\linewidth}{!}{%
\begin{tabular}{llcccc}
\toprule
Camera & Method & Subject Consistency $\uparrow$ & Background Consistency $\uparrow$ & VoL $\uparrow$ & Vividness Degree $\uparrow$ \\
\midrule
\multirow{4}{*}{Static camera}
 & w/o Drift               & 0.9919          & 0.9818          & 883.55          & 0.0424          \\
 & w/o Grounding           & 0.9918          & \textbf{0.9833} & 882.44          & 0.0392          \\
 & w/o Scene-View Consistent Optimization & 0.9881          & 0.9817          & 876.57          & \textbf{0.0450} \\
 & Ours                    & \textbf{0.9924} & 0.9831          & \textbf{892.43} & 0.0423          \\
\midrule
\multirow{4}{*}{Orbit camera}
 & w/o Drift               & 0.9666          & 0.9615          & 704.34          & 0.5113          \\
 & w/o Grounding           & 0.9649          & 0.9604          & 681.16          & 0.5080          \\
 & w/o Scene-View Consistent Optimization & 0.9642          & 0.9611          & 551.33          & 0.5081          \\
 & Ours                    & \textbf{0.9673} & \textbf{0.9622} & \textbf{709.75} & \textbf{0.5117} \\
\bottomrule
\end{tabular}%
}
\end{table}

\begin{figure}[!t]
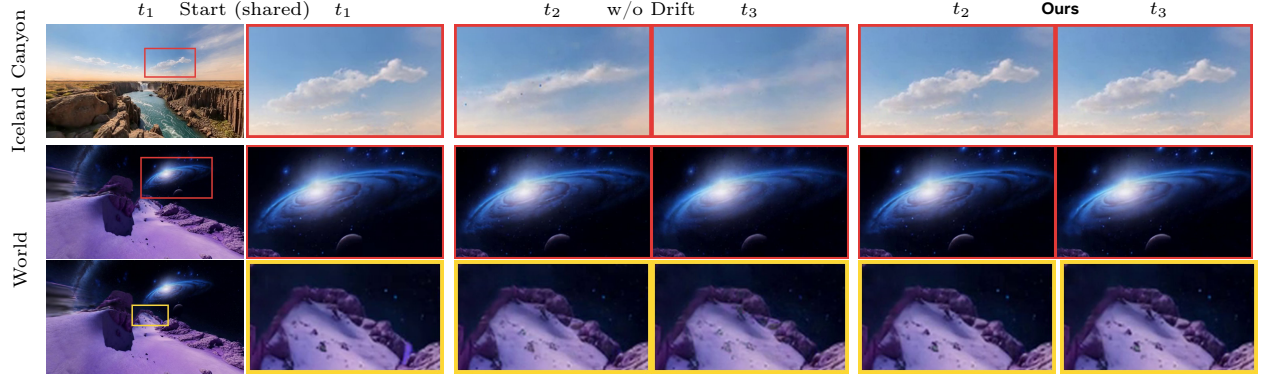

\centering
\ifdefined\driftw\else\newlength{\driftw}\fi
\ifdefined\drifth\else\newlength{\drifth}\fi
\setlength{\driftw}{\dimexpr(\textwidth-1.6em-11pt)/6\relax}
\setlength{\drifth}{0.5714\driftw}   %
\newcommand{\dimg}[1]{\includegraphics[width=\driftw,height=\drifth]{figs/no_drift_demo/#1}}
\newcommand{\dzoom}[4]{\dimg{#1/zoom/#2_t#3_#4}}
\newcommand{\drow}[2]{\dimg{#1/zoom/ours_t1_boxed_#2} & \dzoom{#1}{ours}{1}{#2}
  & \dzoom{#1}{nodrift}{2}{#2} & \dzoom{#1}{nodrift}{3}{#2}
  & \dzoom{#1}{ours}{2}{#2} & \dzoom{#1}{ours}{3}{#2}}
\newcommand{\dscene}[2]{\raisebox{0pt}[0pt][0pt]{\makebox[1.6em]{%
  \rotatebox{90}{\makebox[#1][c]{\scriptsize\shortstack[c]{#2}}}}}}
\newcommand{\dhead}[3]{\parbox[c][2.6ex][c]{2\driftw}{%
  \makebox[\driftw][c]{\scriptsize $t_{#1}$}%
  \makebox[0pt][c]{\scriptsize #2}%
  \makebox[\driftw][c]{\scriptsize $t_{#3}$}}}
\renewcommand{\arraystretch}{0}
\begin{tabular}{@{}c@{\hspace{2pt}}c@{\hspace{1pt}}c@{\hspace{4pt}}c@{}c@{\hspace{4pt}}c@{}c@{}}
& \multicolumn{2}{@{}c@{\hspace{4pt}}}{\dhead{1}{Start (shared)}{1}}
& \multicolumn{2}{@{}c@{\hspace{4pt}}}{\dhead{2}{w/o Drift}{3}}
& \multicolumn{2}{@{}c@{}}{\dhead{2}{\textbf{Ours}}{3}} \\
\dscene{\drifth}{Iceland Canyon} & \drow{iceland_canyon}{cloud} \\[3pt]
& \drow{world}{nebula} \\[1pt]
\dscene{\dimexpr2\drifth+1pt\relax}{World} & \drow{world}{peak} \\
\end{tabular}
\caption{\textbf{Effect of drift.} Without drift, fine details degrade over time: the
clouds in \emph{Iceland Canyon} blur and fade away (top row), the galaxy in
\emph{World} becomes blurry (middle row), and the rocks on the ground vanish
(bottom row). In contrast, our full model preserves these details while
maintaining vivid and coherent motion. The first column shows the full frame
at $t_1$ with the zoomed region outlined; the remaining columns zoom into that
region at $t_1$, and at $t_2$ and $t_3$ for each method.}
\label{figs:no_drift_ablation}
\end{figure}

We ablate three components in \cref{tab:ablation_consistency} and one in the periodic deformation paragraph. Besides Vividness, we report VBench~\cite{huang2024vbench} Subject and Background Consistency~\cite{caron2021emerging,radford2021learning}. We also report the variance of the Laplacian (VoL)~\cite{pech2000diatom} as a measure of sharpness.

\textbf{Drift field.}
We remove the Grounded Drift Field and fit all views with the Periodic Deformation Field $\mathcal{P}$ alone (\cref{figs:drift_flow_demo}a). $\mathcal{P}$ is then forced to average inconsistent multi-view observations. As shown in \cref{tab:ablation_consistency}, this approach lowers VoL and consistency across both cameras, and causes fine details to fade over time (\cref{figs:no_drift_ablation}). This confirms the benefit of modeling per-view variation separately.

\textbf{Grounding.}
We also apply Drift Field $\Delta_v$ to the reference view (\cref{figs:drift_flow_demo}b), removing the case split in \cref{eq:grounded}. Motion can then be explained by either $\mathcal{P}$ or $\Delta_v$, and the more flexible, view-conditioned $\Delta_v$ absorbs it. As shown in \cref{tab:ablation_consistency}, since $\Delta$ is discarded at inference, Vividness drops to the lowest among all variants ($0.0423 \rightarrow 0.0392$ under the static camera). Grounding prevents this by requiring $\mathcal{P}$ alone to reproduce the reference dynamics.

\textbf{Scene-view consistent optimization.}
We replace the objective of \cref{sec:optim} with a naive $\ell_1$ loss on all generated views and timesteps, without diffusion-based LPIPS loss refinement~\cite{liu2025free4d,meng2021sdedit,rombach2022high}. Inpainting inconsistencies~\cite{yu2025trajectorycrafter} are baked into the scene, giving the lowest Subject Consistency and a $22\%$ VoL drop under the orbit camera.

\textbf{Periodic deformation.}
\label{para:aperoidc}
We double the period of the Fourier basis in \cref{eq:fourier} to $2T$, which keeps the capacity unchanged but makes the deformation no longer $T$-periodic, so the loop is no longer guaranteed. Under the static camera, Seam SSIM drops from $0.9965$ to $0.9783$: the last frame of each cycle no longer matches its first, causing a visible jump at the seam. MALF, in contrast, barely changes ($0.0809 \rightarrow 0.0788$), showing that the degradation concentrates at the loop boundary rather than in the overall motion.

\section{Conclusion and Limitation}
\label{sec:conclusion}

We presented OuroWorld, a mask-free framework that brings any static 3DGS scene alive as a seamlessly looping 3D cinemagraph. It distills dynamics from a VLM-guided video generation prior into an Inconsistency-Robust Periodic 4DGS. The resulting scenes exhibit diverse motion and illumination change beyond Eulerian flow, loop exactly by construction, and remain sharp despite the inconsistent generated views. With our ground-truth-free evaluation framework, experiments on both reconstructed and generated scenes show consistent gains over prior 3D cinemagraph methods.

\textbf{Limitations.}
Our dynamics are bounded by the video generation prior, so implausible generations propagate to the final scene, although VLM filtering mitigates this. Moreover, completing all views from a single reference video leads to greater disocclusions and inconsistencies at wider viewpoint ranges. Generating multiple reference videos from distinct viewpoints is a promising remedy.

\clearpage

\crefalias{subsection}{appendix}

\section*{Appendix}
\renewcommand{\thesection}{A}
\renewcommand{\thesubsection}{\thesection.\arabic{subsection}}
\renewcommand{\thefigure}{\thesection.\arabic{figure}}
\renewcommand{\thetable}{\thesection.\arabic{table}}

\setcounter{figure}{0}
\setcounter{table}{0}

\section*{Appendix Overview}
\label{suppl_sec:overview}

This appendix complements the main paper with the following details:
\begin{itemize}
    \item \cref{suppl_sec:dataset_detail} describes how the 39 evaluation scenes are collected from Mip-NeRF 360, HY-World 2.0, Marble, and Lyra 2.0.
    \item \cref{suppl_sec:method_detail} details each stage of our pipeline, including the VLM prompt, the video generation and lifting settings, and all hyperparameters of the Inconsistency-Robust Periodic 4DGS.
    \item \cref{suppl_sec:optim_detail} formulates the scene-view consistent optimization with diffusion-based refinement.
    \item \cref{suppl_sec:baseline_detail} describes the baselines, how their inputs are derived from our pipeline, and their native loop periods.
    \item \cref{suppl_sec:evaluation_detail} provides the full Vividness breakdown and further details on the proposed ground-truth-free metrics.
    \item \cref{suppl_sec:user_study_detail} describes the protocol and interface of our user study.
\end{itemize}
Rendered videos of all methods are provided in the supplementary material.

\subsection{Details on Dataset Collection}
\label{suppl_sec:dataset_detail}

Our benchmark contains 39 static 3DGS scenes: 9 reconstructed from Mip-NeRF 360~\cite{barron2022mip} and 10 generated by each of HY-World 2.0~\cite{hy2026hy}, Marble~\cite{worldlabs2025marble}, and Lyra 2.0~\cite{shen2026lyra}. Each scene is paired with a reference camera pose, from which the reference image $I_{\text{ref}}$ is rendered and around which the evaluation trajectories (\cref{sec:exp_protocol}) are defined.

\paragraph{Scene selection.}
For the generated sources, we select scenes that (a) contain content with an obvious natural dynamic, such as water, foliage, cloth, fire, or light; (b) jointly span indoor, outdoor-natural, architectural, and stylized environments; and (c) include a clean, well-covered region large enough to support the $\pm20^\circ$ orbit camera used in evaluation. The reference pose for each scene is manually selected within this region.

\paragraph{Mip-NeRF 360.}
We use all nine scenes from Mip-NeRF 360 and reconstruct each using the official 3DGS implementation~\cite{kerbl20233d} for 30,000 iterations with its default settings.

\paragraph{HY-World 2.0 and Marble.}
Both services host public galleries of generated worlds. We download ready-made scenes as Gaussian-splat files from the Marble platform and the Hunyuan 3D ``Scene to 3D'' platform and use them as-is without any regeneration or re-optimization.

\paragraph{Lyra 2.0.}
Lyra 2.0 does not host downloadable scenes; it is an image-to-video-to-3DGS pipeline that must be run locally. We therefore generate its scenes with the official code release, taking as input the 15 sample images and captions shipped with the repository, and keep 10 of the resulting scenes. Each sample is a 16:9 image paired with a handwritten caption describing a camera move through a static scene; we use the captions unchanged. Lyra 2.0 operates in two stages: a camera-controlled image-to-video model, fine-tuned from Wan 2.1 I2V-14B~\cite{wan2025wan}, synthesizes an exploration video at $832\times480$ and 16 FPS along a prescribed trajectory, and a feed-forward reconstructor, combining ViPE pose estimation~\cite{huang2025vipe}, Depth Anything 3~\cite{lin2025depth}, and Gaussian prediction, lifts the video into 3DGS. We modify both stages as follows.

\emph{Exploration trajectory.}
The default trajectory, a short zoom-in and zoom-out along the optical axis, produces almost no lateral parallax. The reconstructed scene is then well covered only within a narrow cone around the input view, leaving holes under our $\pm20^\circ$ orbit. We replace it with a dolly-then-orbit trajectory, implemented as an additional preset in the official inference code. The camera first pushes forward into the scene for about 10\,s, and then orbits for about 20\,s around a pivot placed at the depth center of the input view, sweeping to one side, back through the center, and to the other side while keeping the pivot centered in the frame. The forward push places the camera well inside the scene, and the orbit provides a wide lateral baseline on both sides of the original optical axis. Each scene thus yields a 480-frame exploration video (30\,s at 16 FPS).

\emph{Reconstruction.}
On these long, wide-baseline videos, the released feed-forward reconstructor produces blurry scenes with many floaters. We instead combine a 3D foundation model for poses and initial geometry with a conventional per-scene 3DGS optimization:
\begin{enumerate}
    \item \emph{Keyframe selection.} We select 64 keyframes per video. We first pick 48 frames at equal arc length in optical-flow space and move each to the sharpest frame within $\pm3$ frames. Since 2D flow under-weights forward dolly motion, we run VGGT-$\Omega$~\cite{wang2026vggt} once on these frames, interpolate the recovered camera path over all frames, and re-select 64 keyframes at equal 3D arc length, where rotation is converted into an equivalent translation at the median scene depth.
    \item \emph{Initialization.} We run VGGT-$\Omega$ (1B parameters, 512\, px input) on the 64 keyframes to obtain camera intrinsics, extrinsics, and per-pixel depth with confidence. We unproject the depth into a colored point cloud after discarding the 30\% least-confident pixels and pixels straddling depth discontinuities, voxel-downsample it with a voxel size of 0.006 in the normalized scene units of VGGT, and export the poses and points as a COLMAP model.
    \item \emph{Optimization.} From this initialization, we optimize a standard 3DGS with gsplat~\cite{ye2025gsplat} for 30k iterations, using an $\ell_1$ + $0.2\cdot$D-SSIM photometric loss, SH degree 3, and adaptive density control stopped at 2,000 iterations. All 64 keyframes are used for training. When every 8th keyframe is held out, the reconstructions reach a mean novel-view PSNR of 22.1\,dB and SSIM of 0.71.
\end{enumerate}

\subsection{Details on Our Pipeline}
\label{suppl_sec:method_detail}

\paragraph{Input preprocessing.}
The input scenes differ in their spherical-harmonics (SH) degree: HY-World 2.0 and Marble scenes are exported with SH degree 0, whereas Lyra 2.0 (reconstructed as in \cref{suppl_sec:dataset_detail}) and Mip-NeRF 360 scenes use SH degree 3. We convert all scenes to SH degree 3 by zero-initializing the missing higher-order coefficients, so that the canonical color $\mathbf{c}$ shares one parameterization across sources.

\subsubsection{Looping Video Generation}
\label{suppl_sec:looping_video_detail}

\paragraph{Dynamics prompt.}
GPT-5.5~\cite{singh2025openai} receives the reference image $I_{\text{ref}}$ together with the instruction in \cref{suppl_fig:vlm_prompt}. The instruction asks for clearly visible motion that may involve multiple elements rather than a single phenomenon, forbids introducing new content, or letting objects cross the camera. The returned prompt conditions the video generator and is reused as the text input of the baselines (\cref{suppl_sec:baseline_detail}). We generate another video-generation prompt using the same GPT prompt to produce KVD reference videos (\cref{suppl_sec:evaluation_detail}).

\paragraph{Video generation.}
We query Seedance 2.0~\cite{seedance2026seedance} with $I_{\text{ref}}$ rendered at $1280\times720$ as both the first and last frames, thereby encouraging the video to return to its initial state. We generate 10-second videos at 720p and 24 FPS without audio, and set the aspect ratio to adaptive so that the output follows the conditioning image and requires no further cropping or reframing.

\begin{figure}[t]
    \centering
    \fbox{\begin{minipage}{0.95\linewidth}
    \small
    You write the final English prompt for an image-to-video model.

    \smallskip
    Look at the reference image and imagine the whole scene naturally coming alive. Describe plausible and obviously visible motion across the scene wherever it would make the image feel vivid, atmospheric, and lived-in. Multiple elements may move at the same time when that feels natural, and their motion may differ in direction, timing, and intensity. Do not limit the animation to a single object or a single motion phenomenon.

    \smallskip
    The primary motion must remain unmistakably visible at normal playback speed, without requiring frame-by-frame comparison, while preserving the image hierarchy. Being subordinate to the composition does not mean being barely perceptible. Give the primary motion a strong, scene-appropriate amplitude reached gradually. Do not weaken the primary motion with stacked terms such as very light, barely moving, almost imperceptible, minimal, slight, and subtle.

    \smallskip
    If nothing in the scene would naturally move, you can add lighting and shadow changes to make the scene alive.

    \smallskip
    Do not add new things to the scene. Only move the existing things in the scene. Do not let anything pass into the foreground from behind the camera. Do not let anything pass the camera, move behind it, and disappear from the video.

    \smallskip
    Keep the subjects, environment, visual identity, and overall composition recognizable. The result should feel as if the photographed moment has come to life, not like an unrelated scene or a new narrative event. Keep the camera fixed and make the motion flow naturally as a seamless loop.

    \smallskip
    Write one concise, concrete prompt for the video-generation model. Describe what moves and how the scene feels in motion. Output only the final prompt in English, without analysis, headings, bullet points, JSON, Markdown, or quotation marks.
    \end{minipage}}
    \caption{\textbf{Instruction given to GPT-5.5} together with the reference image to write the dynamics prompt.}
    \label{suppl_fig:vlm_prompt}
\end{figure}

\subsubsection{Dynamic Point Cloud Lifting}
\label{suppl_sec:lifting_detail}

\paragraph{Inputs.}
We reconstruct the dynamic point cloud with a single forward pass of VGGT-$\Omega$~\cite{wang2026vggt} on 45 images: 21 static multi-view renders at $t{=}0$ and 24 temporal frames of the reference video. The static renders are produced by the input 3DGS along a purely horizontal orbit, with azimuth $\phi$ sampled uniformly in $[-20^\circ, 20^\circ]$ at $2^\circ$ intervals and zero elevation and radius offset, so the views are ordered monotonically. The central view ($\phi{=}0^\circ$) is the reference view, and the remaining 20 views coincide with the training viewpoints of \cref{suppl_sec:multiview_detail}. For the temporal frames, we sample 25 sparse frames from $V_{\text{ref}}$. Since its first frame is $I_{\text{ref}}$, which is identical to the static render at $\phi{=}0^\circ$, we drop it and feed only the remaining 24 frames, all assigned to the reference view.

\paragraph{Preprocessing.}
Each image is center-cropped to an aspect ratio within $[0.5, 2.0]$, resized so that its longer side is 512 pixels, and aligned to the patch size of 16, e.g., $512\times288$ for a 16:9 input. All 45 images share the same crop and resize, so the reference intrinsics are transformed only once and apply to every input.

\paragraph{Outputs.}
We do not use the camera poses predicted by VGGT-$\Omega$; all cameras are taken from the input 3DGS, and the predicted poses serve only as a sanity check. We remove the 10\% least-confident points of each frame and align the resulting dynamic point cloud to the coordinate frame of the input 3DGS with an affine transformation.

\subsubsection{Multi-View Video Generation}
\label{suppl_sec:multiview_detail}

\paragraph{Cameras.}
For each scene, we manually select a pivot $\mathbf{p}$ by back-projecting a pixel of the reference view. The $N{=}20$ training cameras face $\mathbf{p}$ and rotate within $\pm20^\circ$ about the axis through $\mathbf{p}$ parallel to the yaw axis of the reference camera, at radius $r = \lVert \mathbf{o}_{\text{ref}} - \mathbf{p} \rVert$, where $\mathbf{o}_{\text{ref}}$ is the reference camera center. The orbit camera used for evaluation (\cref{sec:exp_protocol}) shares the same pivot but uses a radius of $0.95r$, thereby probing viewpoints off the training trajectory.

\paragraph{Video inpainting.}
We render the aligned dynamic point cloud from each training camera to obtain incomplete videos, which TrajectoryCrafter~\cite{yu2025trajectorycrafter} completes at $672\times384$ with 50 sampling steps and a guidance scale of 6.0 (seed 43). All target views share the same initial noise, which reduces cross-view appearance drift.

\subsubsection{Inconsistency-Robust Periodic 4DGS}
\label{suppl_sec:irp4dgs_detail}

\paragraph{Architecture.}
The Periodic Deformation Field $\mathcal{P}$ uses a single-scale triplane with a resolution of $64$ per axis. The triplane feature is split into Fourier coefficients with $K{=}4$ frequencies (\cref{eq:fourier}), and the time-dependent feature is decoded by an MLP of width 128 with separate heads for the offsets of position, rotation, scale, and SH coefficients. The Grounded Drift Field $\Delta$ uses the same architecture and heads, and conditions on the time embedding $\gamma(t)$ and the per-view embedding $\mathbf{e}_v$ instead of the periodic feature. A single triplane $\mathbf{F}_\Delta$ is shared across all non-reference views, and views are distinguished only by $\mathbf{e}_v$. We share it because we observe that the inconsistencies produced by TrajectoryCrafter are correlated across views, i.e., independently inpainted views tend to exhibit similar artifacts, so a shared spatial feature explains them with fewer parameters than per-view features.

\paragraph{Hyperparameters.}
\cref{suppl_tab:hyperparams} summarizes the hyperparameters of the full pipeline. The number of optimization iterations is adjusted per scene.

\begin{table}[t]
    \centering
    \caption{\textbf{Hyperparameters of OuroWorld.}}
    \label{suppl_tab:hyperparams}
    \small
    \begin{tabular}{llc}
        \hline
        Stage & Hyperparameter & Value \\
        \hline
        Looping video & VLM / video generator & GPT-5.5 / Seedance 2.0 \\
        & Conditioning image resolution & $1280\times720$ \\
        & Duration / frame rate & 10\,s / 24 FPS \\
        \hline
        Lifting & Static renders / temporal frames & 21 / 24 \\
        & Input resolution & $512\times288$ \\
        & Discarded low-confidence points & 10\% per frame \\
        \hline
        Multi-view video & Number of views $N$ & 20 \\
        & Yaw range & $\pm20^\circ$ \\
        & Inpainting resolution & $672\times384$ \\
        & Sampling steps / guidance scale & 50 / 6.0 \\
        \hline
        IRP-4DGS & Triplane resolution ($\mathcal{P}$ and $\Delta$) & 64 \\
        & Temporal bandwidth $K$ & 4 \\
        & MLP width & 128 \\
        & Loop period $T$ & 10s \\
        & Loss weights $\lambda_1$ / $\lambda_p$ & 1 / 0.2 \\
        & Iterations & per scene \\
        \hline
    \end{tabular}
\end{table}

\subsection{Details on Scene-View Consistent Optimization}
\label{suppl_sec:optim_detail}

We detail the objective of \cref{sec:optim}, which follows the consistent 4D-GS optimization of Free4D~\cite{liu2025free4d}. Let $\hat{I}_{t,v}$ denote the rendering of $\mathcal{G}_{t,v}$ (\cref{eq:grounded}) at time $t$ and view $v$, and let $I_{t,v}$ denote the corresponding supervision frame.

\paragraph{Consistent supervision.}
Two sets of frames are highly consistent with the input scene: the reference video $V_{\text{ref}}$, which comes directly from the video model, and the $t{=}0$ renders of the input 3DGS at all training views, which coincide with the scene at the start of the loop since $V_{\text{ref}}$ begins at $I_{\text{ref}}$. We denote their union by $\mathcal{S}$ and supervise it at the pixel level:
\begin{equation}
    \mathcal{L}_1 = \sum_{(t,v)\in\mathcal{S}} \big\lVert \hat{I}_{t,v} - I_{t,v} \big\rVert_1.
\end{equation}

\paragraph{Diffusion-based refinement.}
The remaining generated frames, $(t,v)\notin\mathcal{S}$, are less reliable, so we do not use them as pixel-level targets. Instead, we use them to guide the refinement of the current rendering. We encode $\hat{I}_{t,v}$ into a latent with the encoder $\mathcal{E}$ of Stable Diffusion v1.5~\cite{rombach2022high}, perturb it with noise and denoise it, similar to SDEdit~\cite{meng2021sdedit}. At each denoising step $i$, the model predicts a clean latent $\hat{\mathbf{z}}_{0|i}$, which we modulate toward the generated frame $\mathbf{z}^{g} = \mathcal{E}(I_{t,v})$:
\begin{equation}
    \tilde{\mathbf{z}}_{0|i} = w_i\,\gamma_i\,\mathbf{z}^{g} + (1 - w_i)\,\hat{\mathbf{z}}_{0|i},
    \qquad
    \gamma_i = \frac{\operatorname{std}(\hat{\mathbf{z}}_{0|i})}{\operatorname{std}(\mathbf{z}^{g})},
\end{equation}
where $w_i \in \{0.5, 0.4, 0.3, 0.2, 0.1\}$ controls the influence of the generated frame, linearly decreasing over the $i = 1, \dots, 5$ denoising steps, and $\gamma_i$ matches the latent statistics to avoid over-exposure. The modulated $\tilde{\mathbf{z}}_{0|i}$ replaces $\hat{\mathbf{z}}_{0|i}$ in the subsequent denoising step, and the final latent is decoded into the refined image $\tilde{I}_{t,v}$. The refined image inherits the dynamics of the generated frame while staying close to the current scene, so the inconsistencies of the generated frame are not transferred at the pixel level. We supervise the rendering with a perceptual loss:
\begin{equation}
    \mathcal{L}_p = \sum_{(t,v)\notin\mathcal{S}} \operatorname{LPIPS}\big(\hat{I}_{t,v}, \tilde{I}_{t,v}\big),
\end{equation}
where $\tilde{I}_{t,v}$ is treated as a constant. For non-reference views, both losses are computed on renderings that include the drift $\Delta_v$, following \cref{eq:grounded}.

\paragraph{Schedule.}
Since the refinement relies on a reasonable rendering, we first optimize with $\mathcal{L}_1$ alone, then with the full objective $\mathcal{L} = \lambda_1\mathcal{L}_1 + \lambda_p\mathcal{L}_p$. The hyperparameters are listed in \cref{suppl_tab:hyperparams}.

\subsection{Details on Baselines}
\label{suppl_sec:baseline_detail}

\subsubsection{Methods}

\paragraph{Gaussians-to-Life}~\cite{wimmer2025gaussians} takes as input a 3DGS, a 3D mask, and a text prompt. It uses a video diffusion model~\cite{blattmann2023stable} as the motion prior, lifts the generated 2D motion to 3D point flow via point tracking~\cite{karaev2024cotracker} and depth estimation, and uses this flow to drive the Gaussians. It does not aim to produce looping motion; for all period-dependent evaluation, we treat the length of its generated motion as its period.

\paragraph{3D Cinemagraphy}~\cite{li20233d} takes as input a single image and a 2D mask. It predicts motion with an Eulerian motion model trained on fluid videos~\cite{holynski2021animating}, lifts the image to a layered depth image (LDI)~\cite{shade1998layered,shih20203d}, and produces a loop by symmetrically splatting the LDI features forward and backward in time.

\paragraph{3D-MOM}~\cite{jin2025optimizing} takes as input a single image and a 2D mask. It back-projects the 2D Eulerian motion into a 4D Gaussian representation~\cite{wu20244d}. Its 2D motion prior loops through symmetric splatting, but the optimized 4D representation is not guaranteed to loop.

\paragraph{LoopGaussian}~\cite{li2024loopgaussian} takes as input a 3DGS and a motion mask. It heuristically moves Gaussians toward positions with similar features and forms a loop by bidirectionally blending an Eulerian motion field. It updates only the Gaussian positions.

For fairness, we also apply the same GPT-5.5 filtering for our method to all baselines.

\subsubsection{Input Alignment}

All baselines require inputs beyond a static 3DGS. We derive them from our pipeline so that the same scene, view, and motion description drives every method. Input images are rendered from the input 3DGS at the reference pose, and text prompts are the GPT-5.5 dynamics prompts of \cref{suppl_sec:looping_video_detail}.

\begin{figure}[t]
    \centering
    \begin{minipage}[t]{0.49\linewidth}
        \centering
        \includegraphics[width=\linewidth]{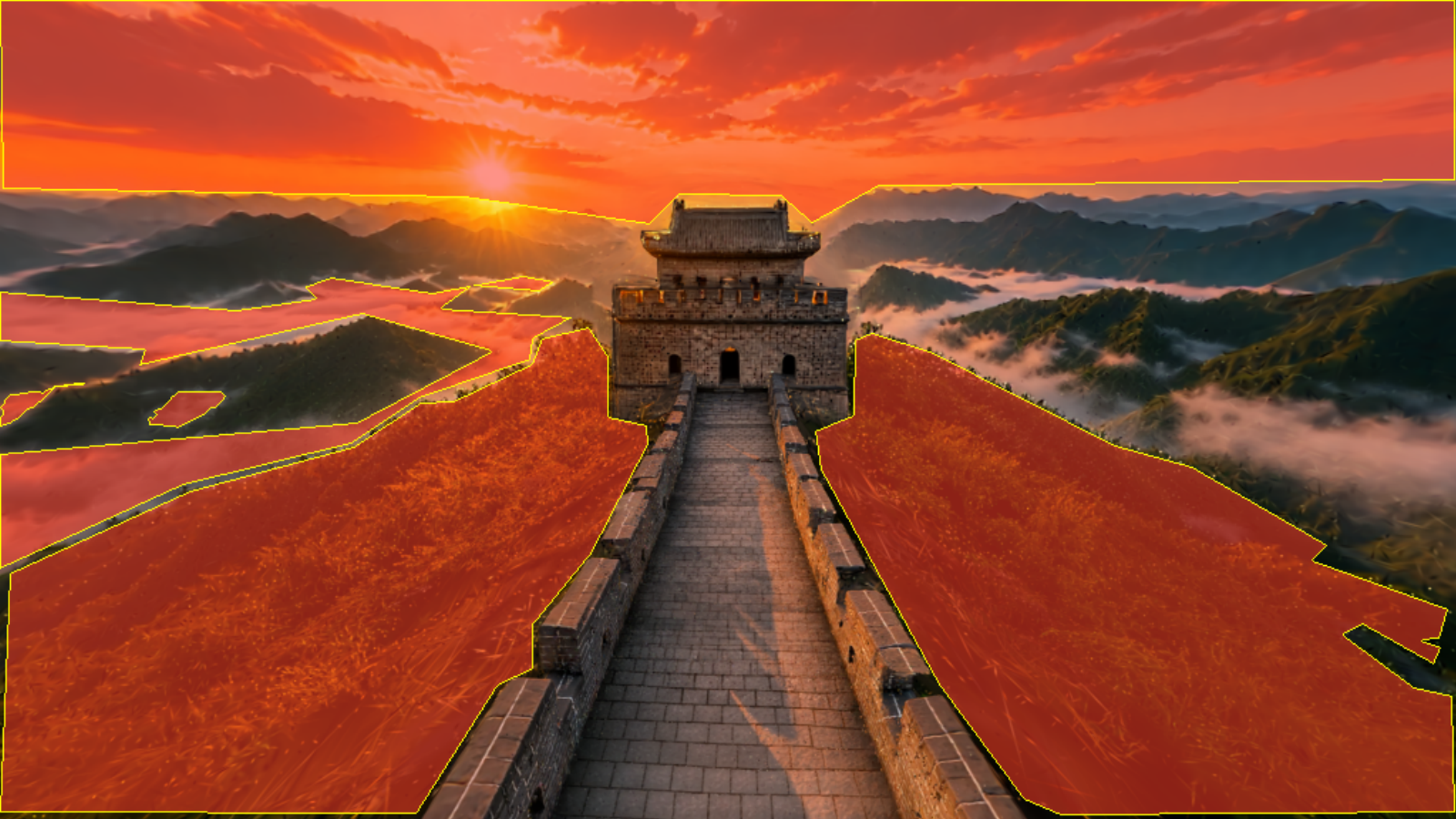}
        \\[0.2em]
        {\small (a) 2D motion mask}
    \end{minipage}\hfill
    \begin{minipage}[t]{0.49\linewidth}
        \centering
        \includegraphics[width=\linewidth]{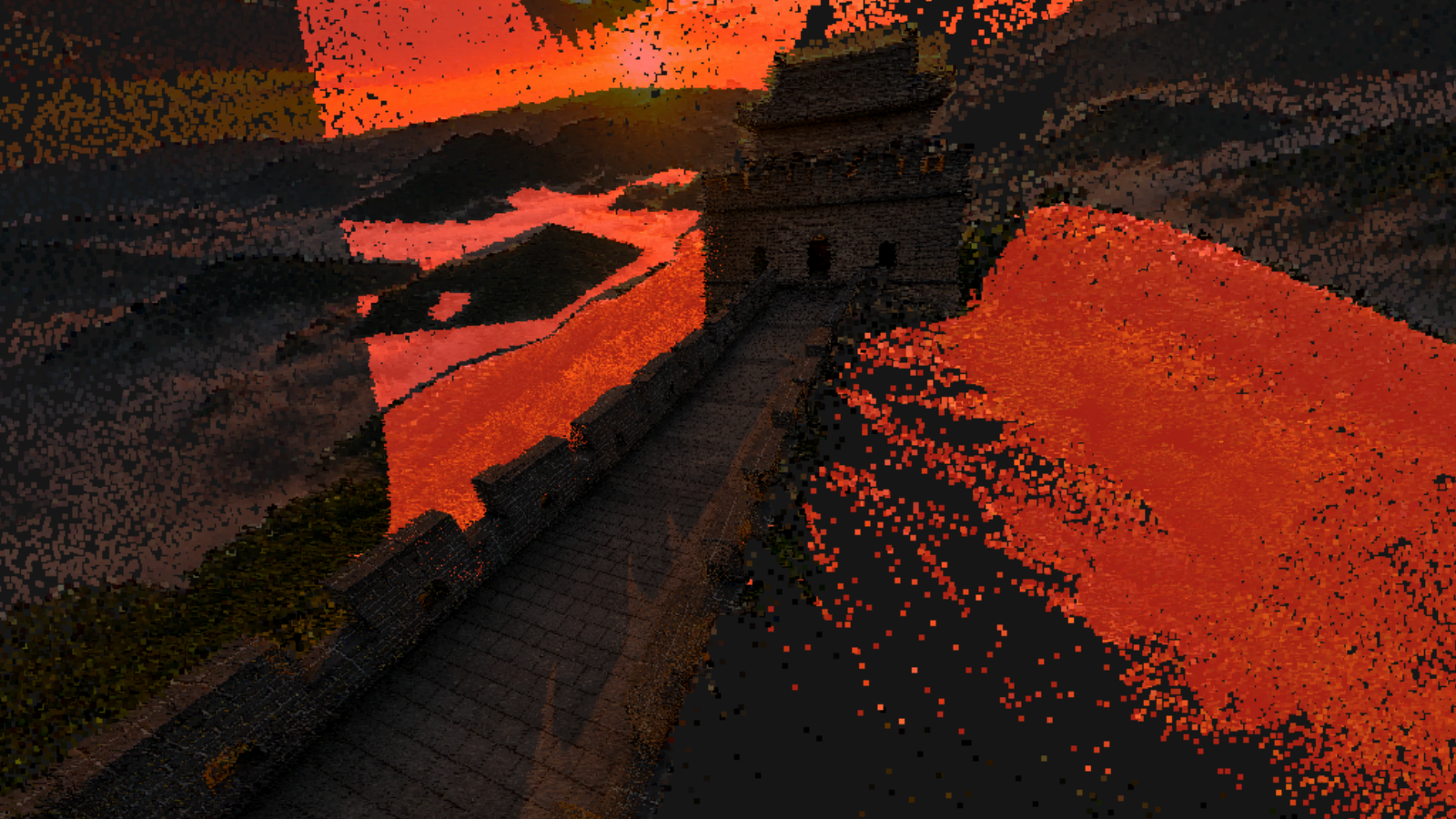}
        \\[0.2em]
        {\small (b) 3D motion mask}
    \end{minipage}
    \caption{\textbf{Motion masks.} (a) The 2D motion mask obtained by segmenting the moving entities in the reference image with SAM~3. (b) The corresponding 3D motion mask.}
    \label{suppl_fig:motion_mask}
\end{figure}

\paragraph{2D motion mask.}
We derive the 2D motion mask from the dynamics prompt rather than from the generated videos, so that it reflects the intended motion instead of any single method's output (\cref{suppl_fig:motion_mask}a). A deterministic, rule-based parser first extracts the moving entities from the prompt, i.e., nouns paired with a motion verb such as \emph{river}, \emph{trees}, or \emph{grass}, together with any entities the prompt explicitly marks as static. Because the parser is rule-based rather than LLM-based, the extracted entities are fully reproducible. Each entity phrase is then segmented separately on the reference image with SAM~3~\cite{carion2026sam} using text-concept segmentation, since short noun phrases give more reliable concept masks than the full prompt. We filter the predictions with the following rules:
(i) we keep instances with a confidence score of at least 0.45 to suppress false positives;
(ii) we discard any instance covering more than 85\% of the frame, which typically indicates a degenerate prediction that selects the whole image;
(iii) we discard any entity whose union covers more than 90\% of the frame, as such a mask cannot separate moving from static regions; and
(iv) we remove connected components smaller than 0.08\% of the image to eliminate speckle noise.
We then subtract the masks of the static entities, which handles cases where a static object overlaps a moving region, such as a bridge spanning a river. The union of the remaining masks forms the 2D motion mask. Finally, every mask is verified by a human before it is used for evaluation.

\paragraph{3D motion mask.}
Among the baselines, only Gaussians-to-Life requires a 3D motion mask; all others use the 2D mask directly. For it, we lift the 2D motion mask to 3D (\cref{suppl_fig:motion_mask}b). We project every Gaussian center into the reference camera and select the Gaussians that fall inside the 2D mask, lie in front of the camera, and have depth within $[d_{\min}, d_{\max}]$. This interval is set automatically from the 5th and 95th percentiles of the in-mask Gaussian depths, padded by 10\% on each side, which excludes background Gaussians that project into the mask from behind the moving entity. As with the 2D mask, the selection is verified by a human; in a typical scene, it contains 14\% of the Gaussians.

\subsubsection{Native Periods}

As described in \cref{sec:exp_protocol}, we keep the native period of each method and repeat it to fill the 20-second evaluation video. \cref{suppl_tab:baseline_summary} summarizes the inputs and native periods of all methods.

\begin{table}[t]
    \centering
    \caption{\textbf{Inputs and native loop periods of all methods.} For Gaussians-to-Life, which does not loop, the period is the length of its generated motion.}
    \label{suppl_tab:baseline_summary}
    \small
    \begin{tabular}{lllc}
        \hline
        Method & Scene input & Motion input & Native period \\
        \hline
        Gaussians-to-Life & 3DGS & 3D mask + text & 0.5\,s \\
        3D Cinemagraphy & Single image & 2D mask & 2.4\,s \\
        3D-MOM & Single image & 2D mask & 2.0\,s \\
        LoopGaussian & 3DGS & 2D mask & 2.0\,s \\
        Ours & 3DGS & None & 10\,s \\
        \hline
    \end{tabular}
\end{table}

\subsection{Details on Evaluation}
\label{suppl_sec:evaluation_detail}

\subsubsection{Vividness}

\paragraph{Thresholds.}
Each rendered video is temporally downsampled from 30 to 3 FPS, and the three components are computed between adjacent frames. The thresholds of Motion Variation and Illumination Variation, i.e., a flow magnitude of $0.3\%\times\min(H,W)$ (about 1.2 pixels at $672\times384$) and a luminance change of 20\%, were chosen empirically. We visually inspected which regions appear to move or change in illumination, tested several candidate thresholds, and selected the pair whose masks best agreed with human perception.

\paragraph{Full results.}
\cref{suppl_tab:vividness_baseline,suppl_tab:vividness_ablation} report each component of the Vividness Degree for the comparison and the ablation, respectively. Under the static camera, Gaussians-to-Life attains the highest Motion and Visual Variation, because it loops over only 0.5\,s, and its low MALF (\cref{tab:quantitative}) shows that this motion does not form a coherent loop. Our method obtains by far the highest Illumination Variation, which reflects the illumination changes that Eulerian baselines cannot express. Under the orbit camera, Motion Variation is dominated by the camera motion. It is nearly identical across methods, so the differences mainly come from Illumination Variation, where our method again leads. In the ablation, removing grounding primarily reduces Motion Variation (0.0413 $\rightarrow$ 0.0324 under the static camera), confirming that ungrounded drift absorbs object motion. Removing scene-view consistent optimization slightly increases the static-camera components, which we attribute to inpainting inconsistencies baked into the scene that flicker over time rather than to meaningful dynamics; this variant also shows the lowest consistency and sharpness in \cref{tab:ablation_consistency}.

\begin{table}[t]
    \centering
    \caption{\textbf{Vividness breakdown against baselines.} MV, IV, and VV denote Motion, Illumination, and Visual Variation; Vividness is their average. Bold denotes the best result.}
    \label{suppl_tab:vividness_baseline}
    \small
    \setlength{\tabcolsep}{4pt}
    \begin{tabular}{lcccccccc}
        \hline
        & \multicolumn{4}{c}{Static camera} & \multicolumn{4}{c}{Orbit camera} \\
        \cline{2-5}\cline{6-9}
        Method & MV $\uparrow$ & IV $\uparrow$ & VV $\uparrow$ & Vividness $\uparrow$ & MV $\uparrow$ & IV $\uparrow$ & VV $\uparrow$ & Vividness $\uparrow$ \\
        \hline
        Gaussians-to-Life & \textbf{0.0588} & 0.0170 & \textbf{0.0364} & 0.0374 & 0.9412 & 0.0837 & 0.5044 & 0.5097 \\
        3D Cinemagraphy & 0.0045 & 0.0069 & 0.0086 & 0.0066 & 0.9348 & 0.0765 & 0.4722 & 0.4945 \\
        LoopGaussian & 0.0021 & 0.0025 & 0.0027 & 0.0024 & 0.9390 & 0.0823 & \textbf{0.5047} & 0.5087 \\
        3D-MOM & 0.0234 & 0.0248 & 0.0259 & 0.0247 & 0.9114 & 0.0970 & 0.4089 & 0.4724 \\
        Ours & 0.0413 & \textbf{0.0518} & 0.0338 & \textbf{0.0423} & \textbf{0.9421} & \textbf{0.1146} & 0.4784 & \textbf{0.5117} \\
        \hline
    \end{tabular}
\end{table}

\begin{table}[t]
    \centering
    \caption{\textbf{Vividness breakdown of the ablation.} SVCO denotes scene-view consistent optimization. Bold denotes the best result.}
    \label{suppl_tab:vividness_ablation}
    \small
    \setlength{\tabcolsep}{4pt}
    \begin{tabular}{lcccccccc}
        \hline
        & \multicolumn{4}{c}{Static camera} & \multicolumn{4}{c}{Orbit camera} \\
        \cline{2-5}\cline{6-9}
        Method & MV $\uparrow$ & IV $\uparrow$ & VV $\uparrow$ & Vividness $\uparrow$ & MV $\uparrow$ & IV $\uparrow$ & VV $\uparrow$ & Vividness $\uparrow$ \\
        \hline
        w/o Drift & 0.0414 & 0.0519 & 0.0340 & 0.0424 & 0.9420 & \textbf{0.1151} & 0.4768 & 0.5113 \\
        w/o Grounding & 0.0324 & \textbf{0.0528} & 0.0323 & 0.0392 & \textbf{0.9421} & 0.1144 & 0.4676 & 0.5080 \\
        w/o SVCO & \textbf{0.0442} & 0.0450 & \textbf{0.0454} & \textbf{0.0450} & 0.9409 & 0.1039 & \textbf{0.4795} & 0.5081 \\
        Ours & 0.0413 & 0.0518 & 0.0338 & 0.0423 & \textbf{0.9421} & 0.1146 & 0.4784 & \textbf{0.5117} \\
        \hline
    \end{tabular}
\end{table}

\subsubsection{Naturalness}
For each scene, we generate 20 videos with MiniMax H3~\cite{minimax2026h3} using its default settings, conditioned on the reference image and the GPT-5.5 dynamics prompt, for a total of 780 videos. For each method, we compute KVD~\cite{unterthiner2018towards} between these videos and the method's 78 rendered videos (39 scenes $\times$ 2 trajectories). Videos are split into non-overlapping 10\,s clips, each sampled to 60 frames at 6\, FPS and $224\times224$, and then encoded into 400-d features using the Kinetics-400 I3D network for FVD. Following KID~\cite{binkowski2018demystifying}, we use the unbiased MMD$^2$ estimator with kernel $k(\mathbf{x},\mathbf{y}) = (\mathbf{x}^\top\mathbf{y}/D + 1)^3$, averaged over 100 random subsets of up to 100 clips. Since static camera videos are rendered from a fixed camera, whereas orbit videos include camera motion, KVD is comparable across methods within each camera setting but not across settings.

\subsubsection{Loop Seam Coherence}

\paragraph{Motion-Aware Loop Fidelity.}
Let $I_t$ be the frame at time $t$ under the static camera and $T$ the native period of the method. We sample $M{=}10$ time points $\{t_i\}_{i=1}^{M}$ uniformly within one period and evaluate \cref{eq:MALF}. The first term, $\operatorname{SSIM}(I_{t_i}, I_{t_i+T})$, measures whether the scene returns to the same state one period later. The second term averages the similarity between $I_{t_i}$ and the other phases in the next cycle, thereby measuring how similar the frames are regardless of phase. Their difference is therefore high only when the motion is substantial and phase-aligned across cycles. A static video has all similarities equal to one and receives $\text{MALF}{=}0$. A video that moves but does not return to its initial state has a first term no larger than the second and also receives a near-zero score. Unlike Seam SSIM, MALF therefore cannot be improved by suppressing motion.

\paragraph{Seam SSIM.}
Seam SSIM is the SSIM between the last frame of one cycle and the first frame of the next, i.e., between $I_{T-1/30}$ and $I_{T}$ at 30 FPS. It directly measures the visible jump at the loop boundary, but is trivially high for near-static videos, which is why we report it together with MALF.

\subsection{Details on User Study}
\label{suppl_sec:user_study_detail}

\begin{figure}[t]
    \centering
    \includegraphics[width=\linewidth]{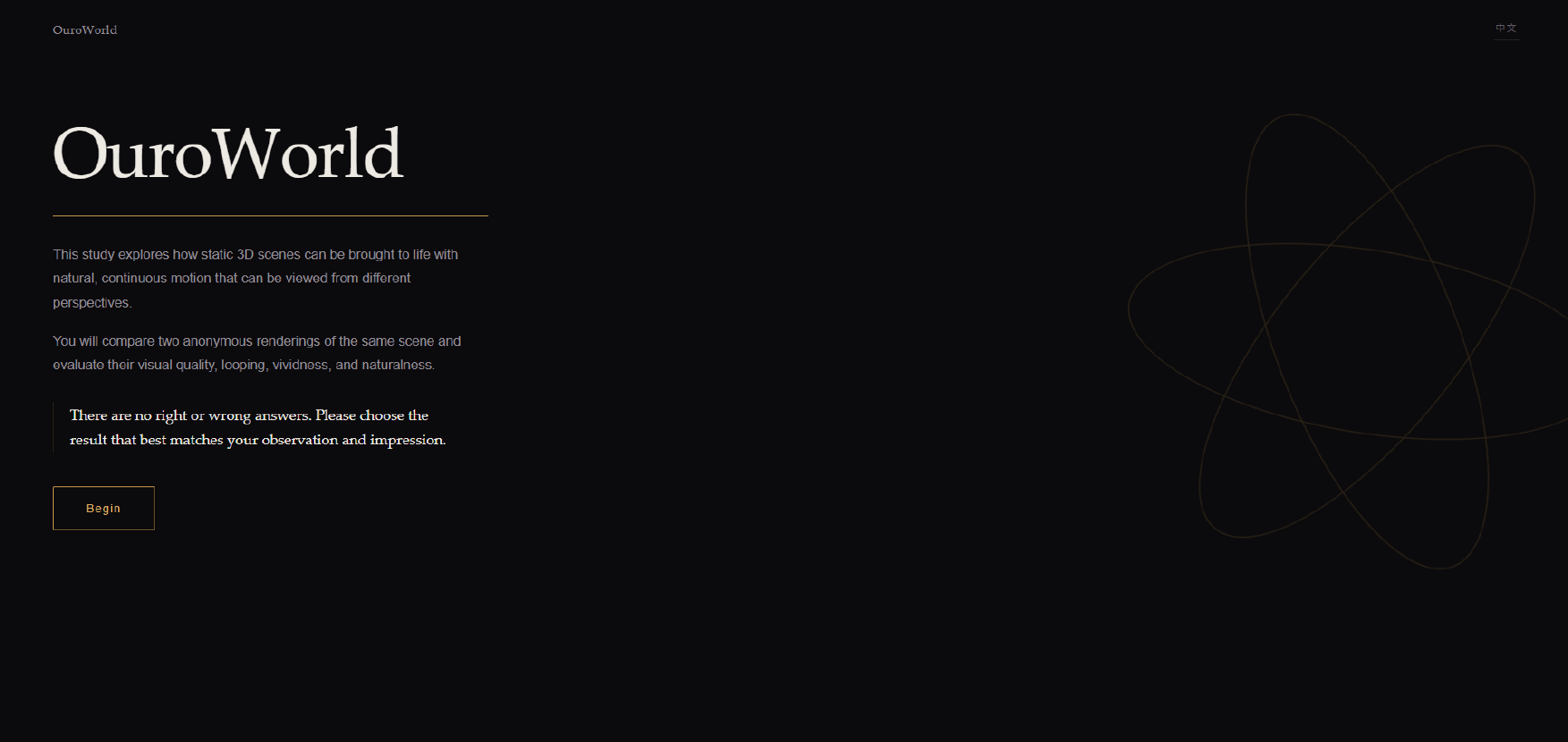}
    \\[0.5em]
    \includegraphics[width=\linewidth]{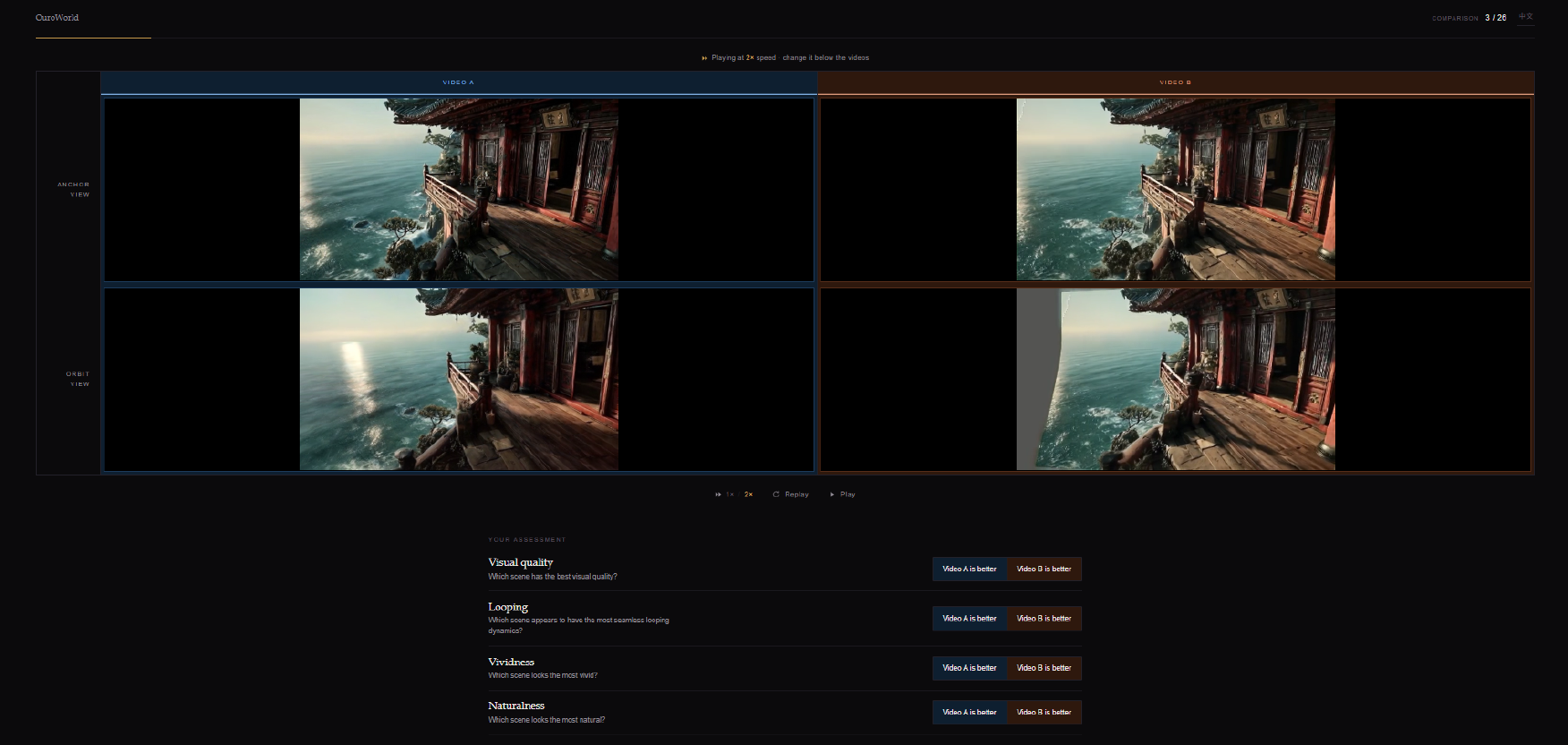}
    \caption{\textbf{User study interface.} Two videos of the same scene are shown side by side in random order (top), and participants choose one for each of the four criteria (bottom).}
    \label{suppl_fig:user_study_ui}
\end{figure}

\paragraph{Protocol.}
We conduct a two-alternative forced-choice study. Each trial shows two rendered videos of the same scene side by side, one from our method and one from a baseline, with the left-right order randomized. Participants answer four questions per trial, one for each criterion:
\begin{itemize}
    \item \emph{Vividness}: Which scene looks the most vivid?
    \item \emph{Naturalness}: Which scene looks the most natural?
    \item \emph{Loop Seam Coherence}: Which scene appears to have the most seamless looping dynamics?
    \item \emph{Visual Quality}: Which scene has the best visual quality?
\end{itemize}
The interface is shown in \cref{suppl_fig:user_study_ui}.

\paragraph{Statistics.}
The study covers all 39 scenes and 4 baselines, i.e., 156 video pairs. Each pair is judged by 5 different participants, and each of the 30 participants judges 26 pairs, resulting in 780 trials in total. Each criterion thus receives 780 votes, 195 per baseline, from which the preference rates in \cref{figs:user_study} are computed. All participants are adult volunteers; participation was voluntary and anonymous, and no personally identifiable information was collected.

\clearpage

\bibliographystyle{abbrv}
\bibliography{main}

\begin{thebibliography}{100}

\bibitem{bahmani2024tc4d}
S.~Bahmani, X.~Liu, W.~Yifan, I.~Skorokhodov, V.~Rong, Z.~Liu, X.~Liu, J.~J.
  Park, S.~Tulyakov, G.~Wetzstein, et~al.
\newblock Tc4d: Trajectory-conditioned text-to-4d generation.
\newblock In {\em European Conference on Computer Vision}, pages 53--72.
  Springer, 2024.

\bibitem{bahmani2026lyra}
S.~Bahmani, T.~Shen, J.~Ren, J.~Huang, Y.~Jiang, H.~Turki, A.~Tagliasacchi,
  D.~Lindell, Z.~Gojcic, S.~Fidler, et~al.
\newblock Lyra: Generative 3d scene reconstruction via video diffusion model
  self-distillation.
\newblock In {\em International Conference on Learning Representations}, volume
  2026, pages 82850--82880, 2026.

\bibitem{bahmani20244d}
S.~Bahmani, I.~Skorokhodov, V.~Rong, G.~Wetzstein, L.~Guibas, P.~Wonka,
  S.~Tulyakov, J.~J. Park, A.~Tagliasacchi, and D.~B. Lindell.
\newblock 4d-fy: Text-to-4d generation using hybrid score distillation
  sampling.
\newblock In {\em 2024 IEEE/CVF Conference on Computer Vision and Pattern
  Recognition (CVPR)}, pages 7996--8006. IEEE, 2024.

\bibitem{bai2012selectively}
J.~Bai, A.~Agarwala, M.~Agrawala, and R.~Ramamoorthi.
\newblock Selectively de-animating video.
\newblock {\em ACM Trans. Graph.}, 31(4):66--1, 2012.

\bibitem{barron2022mip}
J.~T. Barron, B.~Mildenhall, D.~Verbin, P.~P. Srinivasan, and P.~Hedman.
\newblock Mip-nerf 360: Unbounded anti-aliased neural radiance fields.
\newblock In {\em 2022 IEEE/CVF Conference on Computer Vision and Pattern
  Recognition (CVPR)}, pages 5460--5469. IEEE, 2022.

\bibitem{besl1992method}
P.~J. Besl and N.~D. McKay.
\newblock Method for registration of 3-d shapes.
\newblock In {\em Sensor fusion IV: control paradigms and data structures},
  volume 1611, pages 586--606. Spie, 1992.

\bibitem{bi2025mobius}
X.~Bi, J.~Yuan, B.~Liu, Y.~Zhang, X.~Cun, C.-M. Pun, and B.~Xiao.
\newblock Mobius: Text to seamless looping video generation via latent shift.
\newblock In {\em Proceedings of the Special Interest Group on Computer
  Graphics and Interactive Techniques Conference Conference Papers}, pages
  1--10, 2025.

\bibitem{binkowski2018demystifying}
M.~Bi{\'n}kowski, D.~J. Sutherland, M.~Arbel, and A.~Gretton.
\newblock Demystifying mmd gans.
\newblock {\em arXiv preprint arXiv:1801.01401}, 2018.

\bibitem{blattmann2023stable}
A.~Blattmann, T.~Dockhorn, S.~Kulal, D.~Mendelevitch, M.~Kilian, D.~Lorenz,
  Y.~Levi, Z.~English, V.~Voleti, A.~Letts, et~al.
\newblock Stable video diffusion: Scaling latent video diffusion models to
  large datasets.
\newblock {\em arXiv preprint arXiv:2311.15127}, 2023.

\bibitem{cao2023hexplane}
A.~Cao and J.~Johnson.
\newblock Hexplane: A fast representation for dynamic scenes.
\newblock In {\em 2023 IEEE/CVF Conference on Computer Vision and Pattern
  Recognition (CVPR)}, pages 130--141. IEEE, 2023.

\bibitem{carion2026sam}
N.~Carion, L.~Gustafson, Y.-T. Hu, S.~Debnath, R.~Hu, D.~Suris Coll-Vinent,
  C.~Ryali, K.~V. Alwala, H.~Khedr, A.~Huang, et~al.
\newblock Sam 3: Segment anything with concepts.
\newblock In {\em International conference on learning representations}, volume
  2026, pages 138846--138923, 2026.

\bibitem{caron2021emerging}
M.~Caron, H.~Touvron, I.~Misra, H.~J{\'e}gou, J.~Mairal, P.~Bojanowski, and
  A.~Joulin.
\newblock Emerging properties in self-supervised vision transformers.
\newblock In {\em 2021 IEEE/CVF international conference on computer vision
  (ICCV)}, pages 9630--9640. IEEE, 2021.

\bibitem{chan2022efficient}
E.~R. Chan, C.~Z. Lin, M.~A. Chan, K.~Nagano, B.~Pan, S.~De~Mello, O.~Gallo,
  L.~J. Guibas, J.~Tremblay, S.~Khamis, et~al.
\newblock Efficient geometry-aware 3d generative adversarial networks.
\newblock In {\em Proceedings of the IEEE/CVF conference on computer vision and
  pattern recognition}, pages 16123--16133, 2022.

\bibitem{chan2026adagar}
J.~Chan, Z.~Zhao, and Y.-L. Liu.
\newblock Adagar: Adaptive gabor representation for dynamic scene
  reconstruction.
\newblock {\em arXiv preprint arXiv:2601.00796}, 2026.

\bibitem{chen2026pantheon360}
T.-H. Chen, Y.-H. Chen, T.~Tu, J.-Y. Lee, C.-Y. Wu, F.~Lin, H.~Zhang, D.~Paz,
  X.~Huang, Y.~Guo, et~al.
\newblock Pantheon360: Taming digital twin generation via 3d-aware 360
  $\{$$\backslash$deg$\}$ video diffusion.
\newblock {\em arXiv preprint arXiv:2605.25449}, 2026.

\bibitem{chen2026periodic}
Y.~Chen, C.~Gu, J.~Jiang, X.~Zhu, and L.~Zhang.
\newblock Periodic vibration gaussian: Dynamic urban scene reconstruction and
  real-time rendering.
\newblock {\em International Journal of Computer Vision}, 134(3):83, 2026.

\bibitem{cheng2026anigs}
Y.-C. Cheng, C.~Gao, C.~Chen, T.~Li, R.~Shah, A.~Saraf, C.~Kim, L.~Gui,
  A.~Schwing, J.~Kopf, et~al.
\newblock Anigs: Bridging rendering and diffusion prior for 3d scene animation.
\newblock {\em arXiv preprint arXiv:2607.18539}, 2026.

\bibitem{chien2026splannequin}
H.-J. Chien, Y.-C. Huang, C.-H. Wu, W.-L. Chao, and Y.-L. Liu.
\newblock Splannequin: Freezing monocular mannequin-challenge footage with
  dual-detection splatting.
\newblock In {\em 2026 IEEE/CVF Winter Conference on Applications of Computer
  Vision (WACV)}, pages 8028--8040. IEEE, 2026.

\bibitem{choi2024stylecinegan}
J.~Choi, K.~Seo, A.~Ashtari, and J.~Noh.
\newblock Stylecinegan: landscape cinemagraph generation using a pre-trained
  stylegan.
\newblock In {\em 2024 IEEE/CVF Conference on Computer Vision and Pattern
  Recognition (CVPR)}, pages 7872--7881. IEEE, 2024.

\bibitem{chuang2005animating}
Y.-Y. Chuang, D.~B. Goldman, K.~C. Zheng, B.~Curless, D.~H. Salesin, and
  R.~Szeliski.
\newblock Animating pictures with stochastic motion textures.
\newblock In {\em ACM SIGGRAPH 2005 Papers}, pages 853--860. 2005.

\bibitem{chung2023luciddreamer}
J.~Chung, S.~Lee, H.~Nam, J.~Lee, and K.~M. Lee.
\newblock Luciddreamer: Domain-free generation of 3d gaussian splatting scenes.
\newblock {\em arXiv preprint arXiv:2311.13384}, 2023.

\bibitem{dong2026loopy}
H.~Dong, W.~Wang, C.~Li, J.~Lyu, X.~Wang, and D.~Lin.
\newblock Loopy: Seamless video loop generation via anchored looping shift of
  positional embedding.
\newblock {\em arXiv preprint arXiv:2608.23090}, 2026.

\bibitem{duan20244d}
Y.~Duan, F.~Wei, Q.~Dai, Y.~He, W.~Chen, and B.~Chen.
\newblock 4d-rotor gaussian splatting: towards efficient novel view synthesis
  for dynamic scenes.
\newblock In {\em ACM SIGGRAPH 2024 conference papers}, pages 1--11, 2024.

\bibitem{endo2019animating}
Y.~Endo, Y.~Kanamori, and S.~Kuriyama.
\newblock Animating landscape: self-supervised learning of decoupled motion and
  appearance for single-image video synthesis.
\newblock {\em arXiv preprint arXiv:1910.07192}, 2019.

\bibitem{fan2025spectromotion}
C.-D. Fan, C.-W. Chang, Y.-R. Liu, J.-Y. Lee, J.-L. Huang, Y.-C. Tseng, and
  Y.-L. Liu.
\newblock Spectromotion: Dynamic 3d reconstruction of specular scenes.
\newblock In {\em 2025 IEEE/CVF Conference on Computer Vision and Pattern
  Recognition (CVPR)}, pages 21328--21338. IEEE, 2025.

\bibitem{fan2023simulating}
S.~Fan, J.~Piao, C.~Qian, H.~Li, and K.-Y. Lin.
\newblock Simulating fluids in real-world still images.
\newblock In {\em 2023 IEEE/CVF International Conference on Computer Vision
  (ICCV)}, pages 15876--15885. IEEE, 2023.

\bibitem{fridovich2023k}
S.~Fridovich-Keil, G.~Meanti, F.~R. Warburg, B.~Recht, and A.~Kanazawa.
\newblock K-planes: Explicit radiance fields in space, time, and appearance.
\newblock In {\em 2023 IEEE/CVF Conference on Computer Vision and Pattern
  Recognition (CVPR)}, pages 12479--12488. IEEE, 2023.

\bibitem{halperin2021endless}
T.~Halperin, H.~Hakim, O.~Vantzos, G.~Hochman, N.~Benaim, L.~Sassy, M.~Kupchik,
  O.~Bibi, and O.~Fried.
\newblock Endless loops: detecting and animating periodic patterns in still
  images.
\newblock {\em ACM Transactions on graphics (TOG)}, 40(4):1--12, 2021.

\bibitem{he2024videoscore}
X.~He, D.~Jiang, G.~Zhang, M.~Ku, A.~Soni, S.~Siu, H.~Chen, A.~Chandra,
  Z.~Jiang, A.~Arulraj, et~al.
\newblock Videoscore: Building automatic metrics to simulate fine-grained human
  feedback for video generation.
\newblock In {\em Proceedings of the 2024 Conference on Empirical Methods in
  Natural Language Processing}, pages 2105--2123, 2024.

\bibitem{hollein2023text2room}
L.~H{\"o}llein, A.~Cao, A.~Owens, J.~Johnson, and M.~Nie{\ss}ner.
\newblock Text2room: Extracting textured 3d meshes from 2d text-to-image
  models.
\newblock In {\em 2023 IEEE/CVF International Conference on Computer Vision
  (ICCV)}, pages 7875--7886. IEEE, 2023.

\bibitem{hollein2026world}
L.~H{\"o}llein and M.~Nie{\ss}ner.
\newblock World reconstruction from inconsistent views.
\newblock {\em arXiv preprint arXiv:2603.16736}, 2026.

\bibitem{holynski2021animating}
A.~Holynski, B.~L. Curless, S.~M. Seitz, and R.~Szeliski.
\newblock Animating pictures with eulerian motion fields.
\newblock In {\em Proceedings of the IEEE/CVF Conference on Computer Vision and
  Pattern Recognition}, pages 5810--5819, 2021.

\bibitem{hou20253d}
H.-Y. Hou, C.-C. Hsu, Y.-C. Huang, M.-Y. Shen, W.-F. Sun, C.~Sun, C.-C. Chang,
  Y.-L. Liu, and C.-Y. Lee.
\newblock 3d gaussian splatting with grouped uncertainty for unconstrained
  images.
\newblock In {\em ICASSP 2025-2025 IEEE International Conference on Acoustics,
  Speech and Signal Processing (ICASSP)}, pages 1--5. IEEE, 2025.

\bibitem{huang20242d}
B.~Huang, Z.~Yu, A.~Chen, A.~Geiger, and S.~Gao.
\newblock 2d gaussian splatting for geometrically accurate radiance fields.
\newblock In {\em ACM SIGGRAPH 2024 conference papers}, pages 1--11, 2024.

\bibitem{huang2025vipe}
J.~Huang, Q.~Zhou, H.~Rabeti, A.~Korovko, H.~Ling, X.~Ren, T.~Shen, J.~Gao,
  D.~Slepichev, C.-H. Lin, et~al.
\newblock Vipe: Video pose engine for 3d geometric perception.
\newblock {\em arXiv preprint arXiv:2508.10934}, 2025.

\bibitem{huang2025dreamphysics}
T.~Huang, H.~Zhang, Y.~Zeng, Z.~Zhang, H.~Li, W.~Zuo, and R.~W. Lau.
\newblock Dreamphysics: Learning physics-based 3d dynamics with video diffusion
  priors.
\newblock In {\em Proceedings of the AAAI Conference on Artificial
  Intelligence}, volume~39, pages 3733--3741, 2025.

\bibitem{huang2025voyager}
T.~Huang, W.~Zheng, T.~Wang, Y.~Liu, Z.~Wang, J.~Wu, J.~Jiang, H.~Li, R.~Lau,
  W.~Zuo, et~al.
\newblock Voyager: Long-range and world-consistent video diffusion for
  explorable 3d scene generation.
\newblock {\em ACM Transactions on Graphics (TOG)}, 44(6):1--15, 2025.

\bibitem{huang2026vlm}
T.-W. Huang, F.-E. Yang, M.-H. Chen, Y.-Y. Lin, and Y.-L. Liu.
\newblock Vlm-dreamer: Vlm-imagined bi-directional inpainting for single-image
  360 scene generation.
\newblock In {\em 2026 IEEE International Conference on Image Processing
  (ICIP)}, pages 1--6. IEEE, 2026.

\bibitem{huang2024sc}
Y.-H. Huang, Y.-T. Sun, Z.~Yang, X.~Lyu, Y.-P. Cao, and X.~Qi.
\newblock Sc-gs: Sparse-controlled gaussian splatting for editable dynamic
  scenes.
\newblock In {\em 2024 IEEE/CVF Conference on Computer Vision and Pattern
  Recognition (CVPR)}, pages 4220--4230. IEEE, 2024.

\bibitem{huang2024vbench}
Z.~Huang, Y.~He, J.~Yu, F.~Zhang, C.~Si, Y.~Jiang, Y.~Zhang, T.~Wu, Q.~Jin,
  N.~Chanpaisit, et~al.
\newblock Vbench: Comprehensive benchmark suite for video generative models.
\newblock In {\em 2024 IEEE/CVF Conference on Computer Vision and Pattern
  Recognition (CVPR)}, pages 21807--21818. IEEE, 2024.

\bibitem{hy2026hy}
T.~HY-World, C.~Cao, X.~Zuo, Z.~Wang, Y.~Zhang, J.~Wu, Z.~Liu, Y.~Gong, Y.~Liu,
  B.~Yuan, et~al.
\newblock Hy-world 2.0: A multi-modal world model for reconstructing,
  generating, and simulating 3d worlds.
\newblock {\em arXiv preprint arXiv:2604.14268}, 2026.

\bibitem{jhou2015animating}
W.-C. Jhou and W.-H. Cheng.
\newblock Animating still landscape photographs through cloud motion creation.
\newblock {\em IEEE Transactions on Multimedia}, 18(1):4--13, 2015.

\bibitem{jiang2024animate3d}
Y.~Jiang, C.~Yu, C.~Cao, F.~Wang, W.~Hu, and J.~Gao.
\newblock Animate3d: Animating any 3d model with multi-view video diffusion.
\newblock {\em Advances in Neural Information Processing Systems},
  37:125879--125906, 2024.

\bibitem{jin2025optimizing}
I.-H. Jin, H.~Choo, S.-H. Jeong, H.~Park, J.~Kim, O.-j. Kwon, and K.~Kong.
\newblock Optimizing 4d gaussians for dynamic scene video from single landscape
  images.
\newblock In {\em International Conference on Learning Representations}, volume
  2025, pages 80517--80537, 2025.

\bibitem{karaev2024cotracker}
N.~Karaev, I.~Rocco, B.~Graham, N.~Neverova, A.~Vedaldi, and C.~Rupprecht.
\newblock Cotracker: It is better to track together.
\newblock In {\em European conference on computer vision}, pages 18--35.
  Springer, 2024.

\bibitem{kerbl20233d}
B.~Kerbl, G.~Kopanas, T.~Leimk{\"u}hler, G.~Drettakis, et~al.
\newblock 3d gaussian splatting for real-time radiance field rendering.
\newblock {\em ACM Trans. Graph.}, 42(4):139--1, 2023.

\bibitem{kiray2026promptvfx}
M.~Kiray, P.~Uhlenbruck, N.~Navab, and B.~Busam.
\newblock Promptvfx: Text-driven fields for open-world 3d gaussian animation.
\newblock In {\em 2026 International Conference on 3D Vision (3DV)}, pages
  1556--1566. IEEE, 2026.

\bibitem{kong2024hunyuanvideo}
W.~Kong, Q.~Tian, Z.~Zhang, R.~Min, Z.~Dai, J.~Zhou, J.~Xiong, X.~Li, B.~Wu,
  J.~Zhang, et~al.
\newblock Hunyuanvideo: A systematic framework for large video generative
  models.
\newblock {\em arXiv preprint arXiv:2412.03603}, 2024.

\bibitem{kratimenos2024dynmf}
A.~Kratimenos, J.~Lei, and K.~Daniilidis.
\newblock Dynmf: Neural motion factorization for real-time dynamic view
  synthesis with 3d gaussian splatting.
\newblock In {\em European Conference on Computer Vision}, pages 252--269.
  Springer, 2024.

\bibitem{kulhanek2024wildgaussians}
J.~Kulhanek, S.~Peng, Z.~Kukelova, M.~Pollefeys, and T.~Sattler.
\newblock Wildgaussians: 3d gaussian splatting in the wild.
\newblock {\em arXiv preprint arXiv:2407.08447}, 2024.

\bibitem{lee2026skyfall}
J.-Y. Lee, Y.-R. Liu, S.-R. Tsai, W.-C. Chang, C.-H. Wu, J.~Chan, Z.~Zhao,
  C.~H. Lin, and Y.-L. Liu.
\newblock Skyfall-gs: Synthesizing immersive 3d urban scenes from satellite
  imagery.
\newblock In {\em European Conference on Computer Vision}, pages 431--454.
  Springer, 2026.

\bibitem{lee2024vividdream}
Y.-C. Lee, Y.-T. Chen, A.~Wang, T.-H. Liao, B.~Y. Feng, and J.-B. Huang.
\newblock Vividdream: Generating 3d scene with ambient dynamics.
\newblock {\em arXiv preprint arXiv:2405.20334}, 2024.

\bibitem{lei2025mosca}
J.~Lei, Y.~Weng, A.~W. Harley, L.~Guibas, and K.~Daniilidis.
\newblock Mosca: Dynamic gaussian fusion from casual videos via 4d motion
  scaffolds.
\newblock In {\em 2025 IEEE/CVF Conference on Computer Vision and Pattern
  Recognition (CVPR)}, pages 6165--6177. IEEE, 2025.

\bibitem{li2024loopgaussian}
J.~Li, L.~Cheng, Z.~Wang, T.~Mu, and J.~He.
\newblock Loopgaussian: creating 3d cinemagraph with multi-view images via
  eulerian motion field.
\newblock In {\em Proceedings of the 32nd ACM International Conference on
  Multimedia}, pages 476--485, 2024.

\bibitem{li2024genrc}
M.-F. Li, Y.-F. Ku, H.-X. Yen, C.~Liu, Y.-L. Liu, A.~Y. Chen, C.-H. Kuo, and
  M.~Sun.
\newblock Genrc: Generative 3d room completion from sparse image collections.
\newblock In {\em European Conference on Computer Vision}, pages 146--163.
  Springer, 2024.

\bibitem{li20233d}
X.~Li, Z.~Cao, H.~Sun, J.~Zhang, K.~Xian, and G.~Lin.
\newblock 3d cinemagraphy from a single image.
\newblock In {\em 2023 IEEE/CVF Conference on Computer Vision and Pattern
  Recognition (CVPR)}, pages 4595--4605. IEEE, 2023.

\bibitem{li2024spacetime}
Z.~Li, Z.~Chen, Z.~Li, and Y.~Xu.
\newblock Spacetime gaussian feature splatting for real-time dynamic view
  synthesis.
\newblock In {\em 2024 IEEE/CVF Conference on Computer Vision and Pattern
  Recognition (CVPR)}, pages 8508--8520. IEEE, 2024.

\bibitem{li2021neural}
Z.~Li, S.~Niklaus, N.~Snavely, and O.~Wang.
\newblock Neural scene flow fields for space-time view synthesis of dynamic
  scenes.
\newblock In {\em Proceedings of the IEEE/CVF conference on computer vision and
  pattern recognition}, pages 6498--6508, 2021.

\bibitem{li2024generative}
Z.~Li, R.~Tucker, N.~Snavely, and A.~Holynski.
\newblock Generative image dynamics.
\newblock In {\em 2024 IEEE/CVF Conference on Computer Vision and Pattern
  Recognition (CVPR)}, pages 24142--24153. IEEE, 2024.

\bibitem{liang2025wonderland}
H.~Liang, J.~Cao, V.~Goel, G.~Qian, S.~Korolev, D.~Terzopoulos, K.~N.
  Plataniotis, S.~Tulyakov, and J.~Ren.
\newblock Wonderland: Navigating 3d scenes from a single image.
\newblock In {\em 2025 IEEE/CVF Conference on Computer Vision and Pattern
  Recognition (CVPR)}, pages 798--810. IEEE, 2025.

\bibitem{liao2015fast}
J.~Liao, M.~Finch, and H.~Hoppe.
\newblock Fast computation of seamless video loops.
\newblock {\em ACM Transactions on Graphics (TOG)}, 34(6):1--10, 2015.

\bibitem{liao2013automated}
Z.~Liao, N.~Joshi, and H.~Hoppe.
\newblock Automated video looping with progressive dynamism.
\newblock {\em ACM Transactions on Graphics (TOG)}, 32(4):1--10, 2013.

\bibitem{lin2025depth}
H.~Lin, S.~Chen, J.~Liew, D.~Y. Chen, Z.~Li, G.~Shi, J.~Feng, and B.~Kang.
\newblock Depth anything 3: Recovering the visual space from any views.
\newblock {\em arXiv preprint arXiv:2511.10647}, 2025.

\bibitem{lin2024gaussian}
Y.~Lin, Z.~Dai, S.~Zhu, and Y.~Yao.
\newblock Gaussian-flow: 4d reconstruction with dynamic 3d gaussian particle.
\newblock In {\em 2024 IEEE/CVF Conference on Computer Vision and Pattern
  Recognition (CVPR)}, pages 21136--21145. IEEE, 2024.

\bibitem{ling2024align}
H.~Ling, S.~W. Kim, A.~Torralba, S.~Fidler, and K.~Kreis.
\newblock Align your gaussians: Text-to-4d with dynamic 3d gaussians and
  composed diffusion models.
\newblock In {\em 2024 IEEE/CVF Conference on Computer Vision and Pattern
  Recognition (CVPR)}, pages 8576--8588. IEEE, 2024.

\bibitem{liu2024physics3d}
F.~Liu, H.~Wang, S.~Yao, S.~Zhang, J.~Zhou, and Y.~Duan.
\newblock Physics3d: Learning physical properties of 3d gaussians via video
  diffusion.
\newblock {\em arXiv preprint arXiv:2406.04338}, 2024.

\bibitem{liu2025free4d}
T.~Liu, Z.~Huang, Z.~Chen, G.~Wang, S.~Hu, L.~Shen, H.~Sun, Z.~Cao, W.~Li, and
  Z.~Liu.
\newblock Free4d: Tuning-free 4d scene generation with spatial-temporal
  consistency.
\newblock In {\em Proceedings of the IEEE/CVF International Conference on
  Computer Vision}, pages 25571--25582, 2025.

\bibitem{liu2023robust}
Y.-L. Liu, C.~Gao, A.~Meuleman, H.-Y. Tseng, A.~Saraf, C.~Kim, Y.-Y. Chuang,
  J.~Kopf, and J.-B. Huang.
\newblock Robust dynamic radiance fields.
\newblock In {\em 2023 IEEE/CVF Conference on Computer Vision and Pattern
  Recognition (CVPR)}, pages 13--23. IEEE, 2023.

\bibitem{luiten2024dynamic}
J.~Luiten, G.~Kopanas, B.~Leibe, and D.~Ramanan.
\newblock Dynamic 3d gaussians: Tracking by persistent dynamic view synthesis.
\newblock In {\em 2024 International Conference on 3D Vision (3DV)}, pages
  800--809. IEEE, 2024.

\bibitem{ma20233d}
L.~Ma, X.~Li, J.~Liao, and P.~V. Sander.
\newblock 3d video loops from asynchronous input.
\newblock In {\em 2023 IEEE/CVF Conference on Computer Vision and Pattern
  Recognition (CVPR)}, pages 310--320. IEEE, 2023.

\bibitem{mahapatra2022controllable}
A.~Mahapatra and K.~Kulkarni.
\newblock Controllable animation of fluid elements in still images.
\newblock In {\em 2022 IEEE/CVF Conference on Computer Vision and Pattern
  Recognition (CVPR)}, pages 3657--3666. IEEE, 2022.

\bibitem{mahapatra2026dreamloop}
A.~Mahapatra, L.~Mai, C.~Ham, and F.~Liu.
\newblock Dreamloop: Controllable cinemagraph generation from a single
  photograph.
\newblock {\em arXiv preprint arXiv:2601.02646}, 2026.

\bibitem{mahapatra2023text}
A.~Mahapatra, A.~Siarohin, H.-Y. Lee, S.~Tulyakov, and J.-Y. Zhu.
\newblock Text-guided synthesis of eulerian cinemagraphs.
\newblock {\em ACM Transactions on Graphics (TOG)}, 42(6):1--13, 2023.

\bibitem{martin2021nerf}
R.~Martin-Brualla, N.~Radwan, M.~S. Sajjadi, J.~T. Barron, A.~Dosovitskiy, and
  D.~Duckworth.
\newblock Nerf in the wild: Neural radiance fields for unconstrained photo
  collections.
\newblock In {\em 2021 IEEE/CVF conference on computer vision and pattern
  recognition (CVPR)}, pages 7206--7215. IEEE, 2021.

\bibitem{meng2021sdedit}
C.~Meng, Y.~He, Y.~Song, J.~Song, J.~Wu, J.-Y. Zhu, and S.~Ermon.
\newblock Sdedit: Guided image synthesis and editing with stochastic
  differential equations.
\newblock {\em arXiv preprint arXiv:2108.01073}, 2021.

\bibitem{mildenhall2021nerf}
B.~Mildenhall, P.~P. Srinivasan, M.~Tancik, J.~T. Barron, R.~Ramamoorthi, and
  R.~Ng.
\newblock Nerf: Representing scenes as neural radiance fields for view
  synthesis.
\newblock {\em Communications of the ACM}, 65(1):99--106, 2021.

\bibitem{minimax2026h3}
{MiniMax}.
\newblock Minimax h3: An open model breaking the boundaries between tasks and
  modalities.
\newblock \url{https://www.minimax.io/blog/minimax-h3}, July 2026.
\newblock Accessed: 2026-09-26.

\bibitem{niklaus20193d}
S.~Niklaus, L.~Mai, J.~Yang, and F.~Liu.
\newblock 3d ken burns effect from a single image.
\newblock {\em ACM Transactions on Graphics (ToG)}, 38(6):1--15, 2019.

\bibitem{oh2017personalized}
T.-H. Oh, K.~Joo, N.~Joshi, B.~Wang, I.~So~Kweon, and S.~Bing~Kang.
\newblock Personalized cinemagraphs using semantic understanding and
  collaborative learning.
\newblock In {\em Proceedings of the IEEE international conference on computer
  vision}, pages 5160--5169, 2017.

\bibitem{pan2025diff4splat}
P.~Pan, C.~Lin, J.~Zhao, C.~Li, Y.~Lin, H.~Li, H.~Yan, K.~Wen, Y.~Lin, Y.~Yuan,
  et~al.
\newblock Diff4splat: Controllable 4d scene generation with latent dynamic
  reconstruction models.
\newblock {\em arXiv preprint arXiv:2511.00503}, 2025.

\bibitem{park2021nerfies}
K.~Park, U.~Sinha, J.~T. Barron, S.~Bouaziz, D.~B. Goldman, S.~M. Seitz, and
  R.~Martin-Brualla.
\newblock Nerfies: Deformable neural radiance fields.
\newblock In {\em 2021 IEEE/CVF International Conference on Computer Vision
  (ICCV)}, pages 5845--5854. IEEE, 2021.

\bibitem{park2021hypernerf}
K.~Park, U.~Sinha, P.~Hedman, J.~T. Barron, S.~Bouaziz, D.~B. Goldman,
  R.~Martin-Brualla, and S.~M. Seitz.
\newblock Hypernerf: A higher-dimensional representation for topologically
  varying neural radiance fields.
\newblock {\em arXiv preprint arXiv:2106.13228}, 2021.

\bibitem{pech2000diatom}
J.~L. Pech-Pacheco, G.~Crist{\'o}bal, J.~Chamorro-Martinez, and
  J.~Fern{\'a}ndez-Valdivia.
\newblock Diatom autofocusing in brightfield microscopy: a comparative study.
\newblock In {\em Proceedings 15th International Conference on Pattern
  Recognition. ICPR-2000}, volume~3, pages 314--317. IEEE, 2000.

\bibitem{pumarola2021d}
A.~Pumarola, E.~Corona, G.~Pons-Moll, and F.~Moreno-Noguer.
\newblock D-nerf: Neural radiance fields for dynamic scenes.
\newblock In {\em 2021 IEEE/CVF Conference on Computer Vision and Pattern
  Recognition (CVPR)}, pages 10313--10322. IEEE, 2021.

\bibitem{radford2021learning}
A.~Radford, J.~W. Kim, C.~Hallacy, A.~Ramesh, G.~Goh, S.~Agarwal, G.~Sastry,
  A.~Askell, P.~Mishkin, J.~Clark, et~al.
\newblock Learning transferable visual models from natural language
  supervision.
\newblock In {\em International conference on machine learning}, pages
  8748--8763. PmLR, 2021.

\bibitem{ren2023dreamgaussian4d}
J.~Ren, L.~Pan, J.~Tang, C.~Zhang, A.~Cao, G.~Zeng, and Z.~Liu.
\newblock Dreamgaussian4d: Generative 4d gaussian splatting.
\newblock {\em arXiv preprint arXiv:2312.17142}, 2023.

\bibitem{ren2024l4gm}
J.~Ren, K.~Xie, A.~Mirzaei, H.~Liang, X.~Zeng, K.~Kreis, Z.~Liu, A.~Torralba,
  S.~Fidler, S.~W. Kim, et~al.
\newblock L4gm: Large 4d gaussian reconstruction model.
\newblock {\em Advances in Neural Information Processing Systems},
  37:56828--56858, 2024.

\bibitem{rombach2022high}
R.~Rombach, A.~Blattmann, D.~Lorenz, P.~Esser, and B.~Ommer.
\newblock High-resolution image synthesis with latent diffusion models.
\newblock In {\em 2022 IEEE/CVF conference on computer vision and pattern
  recognition (CVPR)}, pages 10674--10685. ieee, 2022.

\bibitem{schodl2023video}
A.~Schodl, R.~Szeliski, D.~H. Salesin, and I.~Essa.
\newblock Video textures.
\newblock In {\em Seminal Graphics Papers: Pushing the Boundaries, Volume 2},
  pages 557--570. 2023.

\bibitem{schuhmann2022laion}
C.~Schuhmann, R.~Beaumont, R.~Vencu, C.~Gordon, R.~Wightman, M.~Cherti,
  T.~Coombes, A.~Katta, C.~Mullis, M.~Wortsman, et~al.
\newblock Laion-5b: An open large-scale dataset for training next generation
  image-text models.
\newblock {\em Advances in neural information processing systems},
  35:25278--25294, 2022.

\bibitem{seedance2026seedance}
T.~Seedance, D.~Chen, L.~Chen, X.~Chen, Y.~Chen, Z.~Chen, Z.~Chen, F.~Cheng,
  T.~Cheng, Y.~Cheng, et~al.
\newblock Seedance 2.0: Advancing video generation for world complexity.
\newblock {\em arXiv preprint arXiv:2604.14148}, 2026.

\bibitem{shade1998layered}
J.~Shade, S.~Gortler, L.-w. He, and R.~Szeliski.
\newblock Layered depth images.
\newblock In {\em Proceedings of the 25th annual conference on Computer
  graphics and interactive techniques}, pages 231--242, 1998.

\bibitem{shen2023make}
L.~Shen, X.~Li, H.~Sun, J.~Peng, K.~Xian, Z.~Cao, and G.~Lin.
\newblock Make-it-4d: Synthesizing a consistent long-term dynamic scene video
  from a single image.
\newblock In {\em Proceedings of the 31st ACM international conference on
  multimedia}, pages 8167--8175, 2023.

\bibitem{shen2026lyra}
T.~Shen, S.~Bahmani, K.~He, S.~G. Srinivasan, T.~Cao, J.~Ren, R.~Li, Z.~Wang,
  N.~Sharp, Z.~Gojcic, et~al.
\newblock Lyra 2.0: Explorable generative 3d worlds.
\newblock {\em arXiv preprint arXiv:2604.13036}, 2026.

\bibitem{shih2025prior}
M.-L. Shih, Y.-H. Chen, Y.-L. Liu, and B.~Curless.
\newblock Prior-enhanced gaussian splatting for dynamic scene reconstruction
  from casual video.
\newblock In {\em Proceedings of the SIGGRAPH Asia 2025 Conference Papers},
  pages 1--13, 2025.

\bibitem{shih2024modeling}
M.-L. Shih, J.-B. Huang, C.~Kim, R.~Shah, J.~Kopf, and C.~Gao.
\newblock Modeling ambient scene dynamics for free-view synthesis.
\newblock In {\em ACM SIGGRAPH 2024 Conference Papers}, pages 1--11, 2024.

\bibitem{shih20203d}
M.-L. Shih, S.-Y. Su, J.~Kopf, and J.-B. Huang.
\newblock 3d photography using context-aware layered depth inpainting.
\newblock In {\em 2020 IEEE/CVF Conference on Computer Vision and Pattern
  Recognition (CVPR)}, pages 8025--8035. IEEE, 2020.

\bibitem{singer2023text}
U.~Singer, S.~Sheynin, A.~Polyak, O.~Ashual, I.~Makarov, F.~Kokkinos, N.~Goyal,
  A.~Vedaldi, D.~Parikh, J.~Johnson, et~al.
\newblock Text-to-4d dynamic scene generation.
\newblock {\em arXiv preprint arXiv:2301.11280}, 2023.

\bibitem{singh2025openai}
A.~Singh, A.~Fry, A.~Perelman, A.~Tart, A.~Ganesh, A.~El-Kishky, A.~McLaughlin,
  A.~Low, A.~Ostrow, A.~Ananthram, et~al.
\newblock Openai gpt-5 system card.
\newblock {\em arXiv preprint arXiv:2601.03267}, 2025.

\bibitem{sun2024dimensionx}
W.~Sun, S.~Chen, F.~Liu, Z.~Chen, Y.~Duan, J.~Zhang, and Y.~Wang.
\newblock Dimensionx: Create any 3d and 4d scenes from a single image with
  controllable video diffusion.
\newblock {\em arXiv preprint arXiv:2411.04928}, 2024.

\bibitem{team2025hunyuanworld}
H.~Team, Z.~Wang, Y.~Liu, J.~Wu, Z.~Gu, H.~Wang, X.~Zuo, T.~Huang, W.~Li,
  S.~Zhang, et~al.
\newblock Hunyuanworld 1.0: Generating immersive, explorable, and interactive
  3d worlds from words or pixels.
\newblock {\em arXiv preprint arXiv:2507.21809}, 2025.

\bibitem{tompkin2011towards}
J.~Tompkin, F.~Pece, K.~Subr, and J.~Kautz.
\newblock Towards moment imagery: Automatic cinemagraphs.
\newblock In {\em 2011 conference for visual media production}, pages 87--93.
  IEEE, 2011.

\bibitem{tucker2020single}
R.~Tucker and N.~Snavely.
\newblock Single-view view synthesis with multiplane images.
\newblock In {\em 2020 IEEE/CVF Conference on Computer Vision and Pattern
  Recognition (CVPR)}, pages 548--557. IEEE, 2020.

\bibitem{tulsiani2018layer}
S.~Tulsiani, R.~Tucker, and N.~Snavely.
\newblock Layer-structured 3d scene inference via view synthesis.
\newblock In {\em European Conference on Computer Vision}, pages 311--327.
  Springer, 2018.

\bibitem{unterthiner2018towards}
T.~Unterthiner, S.~Van~Steenkiste, K.~Kurach, R.~Marinier, M.~Michalski, and
  S.~Gelly.
\newblock Towards accurate generative models of video: A new metric \&
  challenges.
\newblock {\em arXiv preprint arXiv:1812.01717}, 2018.

\bibitem{wan2025wan}
T.~Wan, A.~Wang, B.~Ai, B.~Wen, C.~Mao, C.-W. Xie, D.~Chen, F.~Yu, H.~Zhao,
  J.~Yang, et~al.
\newblock Wan: Open and advanced large-scale video generative models.
\newblock {\em arXiv preprint arXiv:2503.20314}, 2025.

\bibitem{wang20254real}
C.~Wang, P.~Zhuang, T.~D. Ngo, W.~Menapace, A.~Siarohin, M.~Vasilkovsky,
  I.~Skorokhodov, S.~Tulyakov, P.~Wonka, and H.-Y. Lee.
\newblock 4real-video: Learning generalizable photo-realistic 4d video
  diffusion.
\newblock In {\em 2025 IEEE/CVF Conference on Computer Vision and Pattern
  Recognition (CVPR)}, pages 17723--17732. IEEE, 2025.

\bibitem{wang2024loopanimate}
F.~Wang, P.~Liu, H.~Hu, D.~Meng, J.~Su, J.~Xu, Y.~Zhang, X.~Ren, and Z.~Zhang.
\newblock Loopanimate: Loopable salient object animation.
\newblock In {\em Proceedings of the 6th ACM International Conference on
  Multimedia in Asia}, pages 1--8, 2024.

\bibitem{wang2025vggt}
J.~Wang, M.~Chen, N.~Karaev, A.~Vedaldi, C.~Rupprecht, and D.~Novotny.
\newblock Vggt: Visual geometry grounded transformer.
\newblock In {\em 2025 IEEE/CVF Conference on Computer Vision and Pattern
  Recognition (CVPR)}, pages 5294--5306. IEEE, 2025.

\bibitem{wang2026vggt}
J.~Wang, M.~Chen, S.~Zhang, N.~Karaev, J.~Sch{\"o}nberger, P.~Labatut,
  P.~Bojanowski, D.~Novotny, A.~Vedaldi, and C.~Rupprecht.
\newblock Vggt-$\omega$.
\newblock {\em arXiv preprint arXiv:2605.15195}, 2026.

\bibitem{wang2022fourier}
L.~Wang, J.~Zhang, X.~Liu, F.~Zhao, Y.~Zhang, Y.~Zhang, M.~Wu, J.~Yu, and
  L.~Xu.
\newblock Fourier plenoctrees for dynamic radiance field rendering in
  real-time.
\newblock In {\em Proceedings of the IEEE/CVF Conference on Computer Vision and
  Pattern Recognition}, pages 13524--13534, 2022.

\bibitem{wang2025shape}
Q.~Wang, V.~Ye, H.~Gao, W.~Zeng, J.~Austin, Z.~Li, and A.~Kanazawa.
\newblock Shape of motion: 4d reconstruction from a single video.
\newblock In {\em 2025 IEEE/CVF International Conference on Computer Vision
  (ICCV)}, pages 9660--9672. IEEE, 2025.

\bibitem{wang2024dust3r}
S.~Wang, V.~Leroy, Y.~Cabon, B.~Chidlovskii, and J.~Revaud.
\newblock Dust3r: Geometric 3d vision made easy.
\newblock In {\em 2024 IEEE/CVF Conference on Computer Vision and Pattern
  Recognition (CVPR)}, pages 20697--20709. IEEE, 2024.

\bibitem{wang2024internvid}
Y.~Wang, Y.~He, Y.~Li, K.~Li, J.~Yu, X.~Ma, X.~Li, G.~Chen, X.~Chen, Y.~Wang,
  et~al.
\newblock Internvid: A large-scale video-text dataset for multimodal
  understanding and generation.
\newblock In {\em International Conference on Learning Representations}, volume
  2024, pages 42055--42079, 2024.

\bibitem{wang2024sea}
Y.~Wang, L.~Lipson, and J.~Deng.
\newblock Sea-raft: Simple, efficient, accurate raft for optical flow.
\newblock In {\em European Conference on Computer Vision}, pages 36--54.
  Springer, 2024.

\bibitem{wang2004image}
Z.~Wang, A.~C. Bovik, H.~R. Sheikh, and E.~P. Simoncelli.
\newblock Image quality assessment: from error visibility to structural
  similarity.
\newblock {\em IEEE transactions on image processing}, 13(4):600--612, 2004.

\bibitem{wiles2020synsin}
O.~Wiles, G.~Gkioxari, R.~Szeliski, and J.~Johnson.
\newblock Synsin: End-to-end view synthesis from a single image.
\newblock In {\em 2020 IEEE/CVF Conference on Computer Vision and Pattern
  Recognition (CVPR)}, pages 7465--7475. IEEE, 2020.

\bibitem{wimmer2025gaussians}
T.~Wimmer, M.~Oechsle, M.~Niemeyer, and F.~Tombari.
\newblock Gaussians-to-life: Text-driven animation of 3d gaussian splatting
  scenes.
\newblock In {\em 2025 International Conference on 3D Vision (3DV)}, pages
  958--968. IEEE, 2025.

\bibitem{worldlabs2025marble}
{World Labs}.
\newblock Marble: A multimodal world model.
\newblock \url{https://www.worldlabs.ai/blog/marble-world-model}, Nov. 2025.
\newblock Official blog post.

\bibitem{wu20244d}
G.~Wu, T.~Yi, J.~Fang, L.~Xie, X.~Zhang, W.~Wei, W.~Liu, Q.~Tian, and X.~Wang.
\newblock 4d gaussian splatting for real-time dynamic scene rendering.
\newblock In {\em 2024 IEEE/CVF Conference on Computer Vision and Pattern
  Recognition (CVPR)}, pages 20310--20320. IEEE, 2024.

\bibitem{wu2025cat4d}
R.~Wu, R.~Gao, B.~Poole, A.~Trevithick, C.~Zheng, J.~T. Barron, and
  A.~Holynski.
\newblock Cat4d: Create anything in 4d with multi-view video diffusion models.
\newblock In {\em Proceedings of the IEEE/CVF Conference on Computer Vision and
  Pattern Recognition}, pages 26057--26068, 2025.

\bibitem{xie2024physgaussian}
T.~Xie, Z.~Zong, Y.~Qiu, X.~Li, Y.~Feng, Y.~Yang, and C.~Jiang.
\newblock Physgaussian: Physics-integrated 3d gaussians for generative
  dynamics.
\newblock In {\em 2024 IEEE/CVF Conference on Computer Vision and Pattern
  Recognition (CVPR)}, pages 4389--4398. IEEE, 2024.

\bibitem{xie2025sv4d}
Y.~Xie, C.-H. Yao, V.~Voleti, H.~Jiang, and V.~Jampani.
\newblock Sv4d: Dynamic 3d content generation with multi-frame and multi-view
  consistency.
\newblock In {\em International Conference on Learning Representations}, volume
  2025, pages 33421--33441, 2025.

\bibitem{xing2024dynamicrafter}
J.~Xing, M.~Xia, Y.~Zhang, H.~Chen, W.~Yu, H.~Liu, G.~Liu, X.~Wang, Y.~Shan,
  and T.-T. Wong.
\newblock Dynamicrafter: Animating open-domain images with video diffusion
  priors.
\newblock In {\em European Conference on Computer Vision}, pages 399--417.
  Springer, 2024.

\bibitem{yang2026ic}
H.-B. Yang, M.-Z. He, C.-W. Yang, J.-C. Chiang, Y.-L. Liu, and W.-H. Peng.
\newblock Ic-4dgs: Illumination-compensated 4d gaussian splatting under
  photometric variations.
\newblock In {\em 2026 IEEE International Conference on Image Processing
  (ICIP)}, pages 1--6. IEEE, 2026.

\bibitem{yang2024deformable}
Z.~Yang, X.~Gao, W.~Zhou, S.~Jiao, Y.~Zhang, and X.~Jin.
\newblock Deformable 3d gaussians for high-fidelity monocular dynamic scene
  reconstruction.
\newblock In {\em 2024 IEEE/CVF Conference on Computer Vision and Pattern
  Recognition (CVPR)}, pages 20331--20341. IEEE, 2024.

\bibitem{yang2024real}
Z.~Yang, H.~Yang, Z.~Pan, and L.~Zhang.
\newblock Real-time photorealistic dynamic scene representation and rendering
  with 4d gaussian splatting.
\newblock In {\em International Conference on Learning Representations}, volume
  2024, pages 9142--9159, 2024.

\bibitem{ye2025gsplat}
V.~Ye, R.~Li, J.~Kerr, M.~Turkulainen, B.~Yi, Z.~Pan, O.~Seiskari, J.~Ye,
  J.~Hu, M.~Tancik, et~al.
\newblock gsplat: An open-source library for gaussian splatting.
\newblock {\em Journal of Machine Learning Research}, 26(34):1--17, 2025.

\bibitem{yeh2012approach}
M.-C. Yeh and P.-Y. Li.
\newblock An approach to automatic creation of cinemagraphs.
\newblock In {\em Proceedings of the 20th ACM international conference on
  Multimedia}, pages 1153--1156, 2012.

\bibitem{yu20244real}
H.~Yu, C.~Wang, P.~Zhuang, W.~Menapace, A.~Siarohin, J.~Cao, L.~A. Jeni,
  S.~Tulyakov, and H.-Y. Lee.
\newblock 4real: Towards photorealistic 4d scene generation via video diffusion
  models.
\newblock {\em Advances in Neural Information Processing Systems},
  37:45256--45280, 2024.

\bibitem{yu2025wonderworld}
H.-X. Yu, H.~Duan, C.~Herrmann, W.~T. Freeman, and J.~Wu.
\newblock Wonderworld: Interactive 3d scene generation from a single image.
\newblock In {\em 2025 IEEE/CVF Conference on Computer Vision and Pattern
  Recognition (CVPR)}, pages 5916--5926. IEEE, 2025.

\bibitem{yu2024wonderjourney}
H.-X. Yu, H.~Duan, J.~Hur, K.~Sargent, M.~Rubinstein, W.~T. Freeman, F.~Cole,
  D.~Sun, N.~Snavely, J.~Wu, et~al.
\newblock Wonderjourney: Going from anywhere to everywhere.
\newblock In {\em 2024 IEEE/CVF Conference on Computer Vision and Pattern
  Recognition (CVPR)}, pages 6658--6667. IEEE, 2024.

\bibitem{yu2025trajectorycrafter}
M.~Yu, W.~Hu, J.~Xing, and Y.~Shan.
\newblock Trajectorycrafter: Redirecting camera trajectory for monocular videos
  via diffusion models.
\newblock In {\em 2025 IEEE/CVF International Conference on Computer Vision
  (ICCV)}, pages 100--111. IEEE, 2025.

\bibitem{yu2024mip}
Z.~Yu, A.~Chen, B.~Huang, T.~Sattler, and A.~Geiger.
\newblock Mip-splatting: Alias-free 3d gaussian splatting.
\newblock In {\em 2024 IEEE/CVF Conference on Computer Vision and Pattern
  Recognition (CVPR)}, pages 19447--19456. IEEE, 2024.

\bibitem{zhang2025monst3r}
J.~Zhang, C.~Herrmann, J.~Hur, V.~Jampani, F.~Cole, D.~Sun, M.-H. Yang, et~al.
\newblock Monst3r: A simple approach for estimating geometry in the presence of
  motion.
\newblock In {\em International Conference on Learning Representations}, volume
  2025, pages 82863--82886, 2025.

\bibitem{zhang2018unreasonable}
R.~Zhang, P.~Isola, A.~A. Efros, E.~Shechtman, and O.~Wang.
\newblock The unreasonable effectiveness of deep features as a perceptual
  metric.
\newblock In {\em 2018 IEEE/CVF conference on computer vision and pattern
  recognition}, pages 586--595. IEEE, 2018.

\bibitem{zhang2024physdreamer}
T.~Zhang, H.-X. Yu, R.~Wu, B.~Y. Feng, C.~Zheng, N.~Snavely, J.~Wu, and W.~T.
  Freeman.
\newblock Physdreamer: Physics-based interaction with 3d objects via video
  generation.
\newblock In {\em European Conference on Computer Vision}, pages 388--406.
  Springer, 2024.

\bibitem{zhang2026fast}
Z.~Zhang, Z.~Lu, P.~Li, H.~Duan, Y.~Liu, P.~Luo, P.~Zhuang, X.~Li, and S.~Ma.
\newblock Fast-gs: Frequency aware space-time gaussian splatting for
  photorealistic dynamic novel view synthesis.
\newblock In {\em ICASSP 2026-2026 IEEE International Conference on Acoustics,
  Speech and Signal Processing (ICASSP)}, pages 9297--9301. IEEE, 2026.

\bibitem{zhao2025genxd}
Y.~Zhao, C.-C. Lin, K.~Lin, Z.~Yan, L.~Li, Z.~Yang, J.~Wang, G.~H. Lee, and
  L.~Wang.
\newblock Genxd: Generating any 3d and 4d scenes.
\newblock In {\em International Conference on Learning Representations}, volume
  2025, pages 95479--95499, 2025.

\bibitem{zhou2024dreamscene360}
S.~Zhou, Z.~Fan, D.~Xu, H.~Chang, P.~Chari, T.~Bharadwaj, S.~You, Z.~Wang, and
  A.~Kadambi.
\newblock Dreamscene360: Unconstrained text-to-3d scene generation with
  panoramic gaussian splatting.
\newblock In {\em European Conference on Computer Vision}, pages 324--342.
  Springer, 2024.

\bibitem{zhou2018stereo}
T.~Zhou, R.~Tucker, J.~Flynn, G.~Fyffe, and N.~Snavely.
\newblock Stereo magnification: Learning view synthesis using multiplane
  images.
\newblock {\em arXiv preprint arXiv:1805.09817}, 2018.

\end{thebibliography}

\end{document}